\pdfoutput=1
\documentclass[11pt]{article}

\usepackage{naacl2021} % final version

\usepackage{times}
\usepackage{latexsym}
\usepackage{graphicx}
\usepackage{subcaption}
\usepackage{latexsym}
\usepackage{lscape}
\usepackage{amsmath}
\usepackage{booktabs}
\DeclareMathOperator*{\argmax}{arg\,max}

\usepackage{amssymb}

\definecolor{britishracinggreen}{rgb}{0.0, 0.26, 0.15}
\newcommand{\dataset}{\textsc{SpartQA}}
\newcommand{\auto}{\textsc{SpartQA-Auto}}
\newcommand{\human}{\textsc{SpartQA-Human}}
\newcommand{\babi}{\text{bAbI}}
\newcommand{\boolq}{\text{boolQ}}
\newcommand{\qtype}{\textsc{Q-Type}}
\usepackage[]{xcolor}
\usepackage{etoolbox}
\usepackage{multirow}
\usepackage{paralist}
\usepackage{soul}
\usepackage{hyperref}

\newrobustcmd{\showcb}[3]{% make comment block
  \csdef{#1}##1{%
    \par\noindent\textcolor{#3}{// \colorbox{#3}{\textcolor{white}{#2}}: ##1}\par%
  }
}
\newrobustcmd{\showcs}[3]{% make comment span
  \csdef{#1}##1{%
    \textcolor{#3}{/* \colorbox{#3}{\textcolor{white}{#2}}: ##1 */}%
  }%
}
\newrobustcmd{\hidec}[1]{%
  \csdef{#1}##1{}
}

\showcs{done}{done}{brown}
\showcs{outline}{outline}{britishracinggreen}

\usepackage{latexsym}

\usepackage[T1]{fontenc}
\usepackage[utf8]{inputenc}

\usepackage{microtype}

\title{\dataset: A Textual Question Answering Benchmark\\ for Spatial Reasoning}

\author{Roshanak Mirzaee$^\bigstar$ ~ Hossein Rajaby Faghihi$^\bigstar$  ~ Qiang Ning$^\spadesuit$\thanks{~~Work was done while at the Allen Institute for AI.} ~Parisa Kordjamshidi$^\bigstar$ ~ \\
  $^\bigstar$Michigan State University \; $^\spadesuit$Amazon \\
  {\tt \{mirzaeem,rajabyfa,kordjams\}@msu.edu \; qning@amazon.com}}

\begin{document}
\maketitle
\begin{abstract}
% This paper proposes \pk{the: a} \pk{first human-generated: i would remove this} question-answering (QA) benchmark for spatial reasoning on natural language text \pk{, where state-of-the-art language models (LM) perform poorly. : JUST A SUGGESTION: . The benchmark contains complex spatial expressions which make the state-of-the-art language models (LM) -tuned on QA- perform poorly.}
% % As our annotation process indicates that human annotations are costly, w
% \pk{A distant supervision method is proposed to improve on this task.: REMOVE}
% We design a grammar \pk{to extend the human-created part of the benchmark and} generate spatial descriptions of a visual scene \pk{automatically.} \pk{, and: We use} a set of spatial reasoning rules \pk{as a source of distant supervision} to \pk{create QA pairs: find the answers to generated questions.} \pk{that can be used as distant supervision signals for spatial reasoning over natural language.: REMOVE} 
% Experiments show that further pretraining LMs on these \pk{automatically generated} QA pairs significantly improves LMs' spatial understanding, which in turn helps to better solve two external datasets, bAbI and boolQ. We believe that this work can foster investigations into more sophisticated models for spatial reasoning over text.

This paper proposes a question-answering (QA) benchmark for spatial reasoning on natural language text which contains more realistic spatial phenomena not covered by prior work and is challenging for state-of-the-art language models (LM). We propose a distant supervision method to improve on this task. Specifically, we design grammar and reasoning rules to automatically generate a spatial description of visual scenes and corresponding QA pairs. Experiments show that further pretraining LMs on these automatically generated data significantly improves LMs' capability on spatial understanding, which in turn helps to better solve two external datasets, bAbI, and boolQ. We hope that this work can foster investigations into more sophisticated models for spatial reasoning over text.
\end{abstract}

%!TEX root=naacl2021.tex
\section{Introduction}
\label{sec:intro}

\begin{figure*}[h]
	\centering
	\begin{subfigure}[b]{\linewidth}%{0.9\linewidth} %QN: I changed them to full width because it looked weird
		\includegraphics[width=\linewidth]{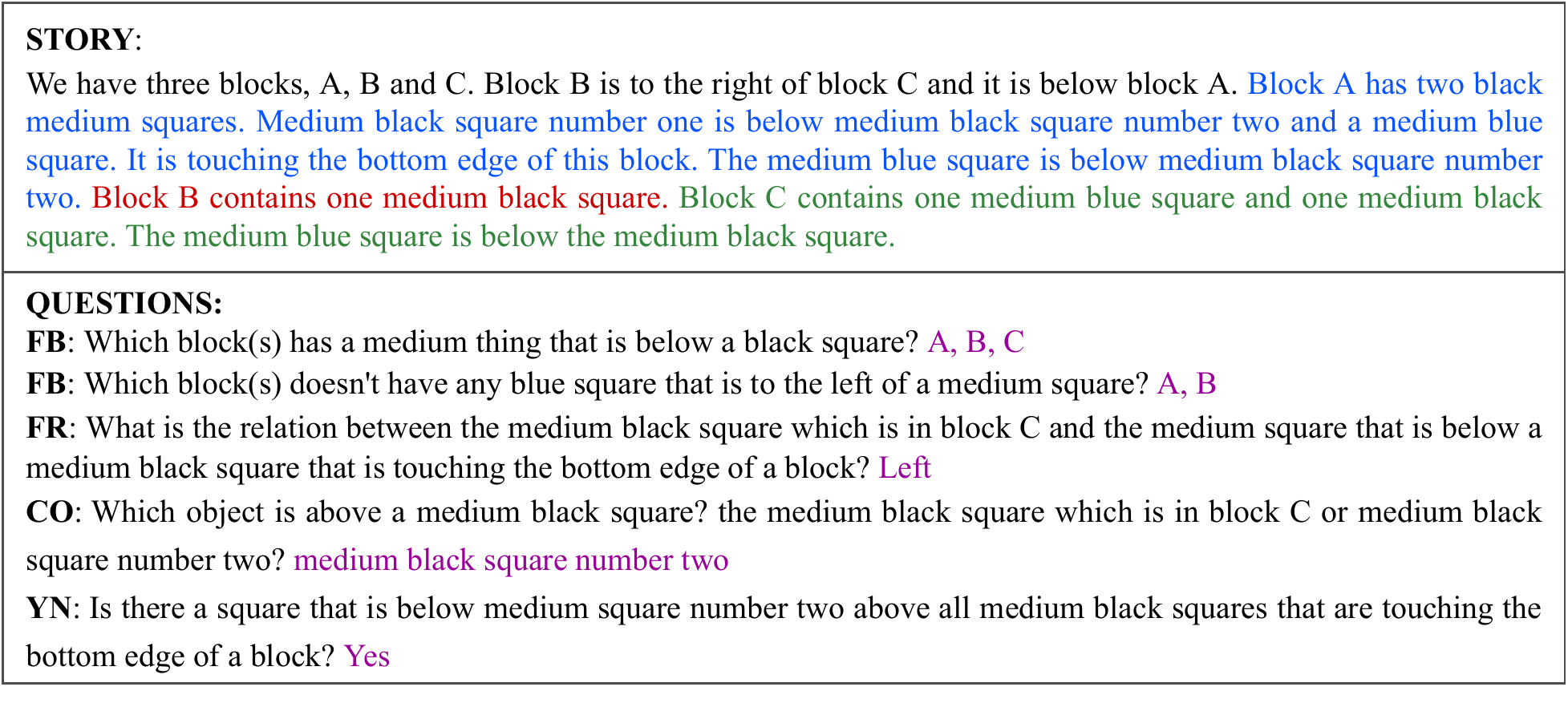}
		\caption{An example story and corresponding questions and answers.
% 		An automatically generated story and corresponding questions and answers.
		}
		\label{fig:story}
	\end{subfigure}
%	\hfill
	\begin{subfigure}[b]{\linewidth}%{0.75\linewidth}
			\includegraphics[width=\linewidth]{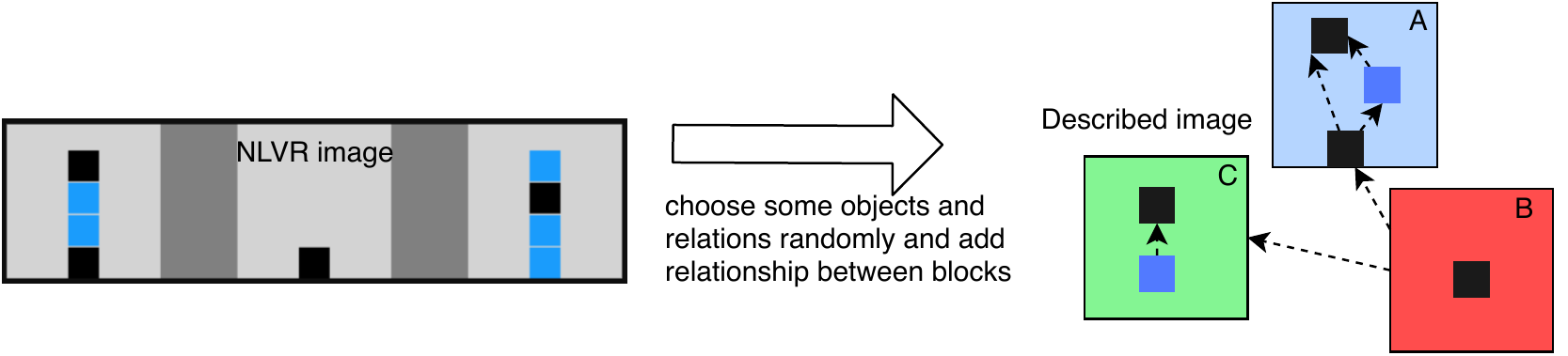}
			\caption{An example NLVR image and the scene created in Fig.~\ref{fig:story}, where the blocks in the NLVR image are rearranged.
% 			The final scene created by the random choices described in Fig \ref{fig:story}.
			}
			\label{fig:nlvr}
    \end{subfigure}
 	\caption{Example from \textsc{SpartQA}~(specifically from \auto)}
	\label{fig:dataset_samples}
\end{figure*}
% \begin{figure*}[hbt!]
%     \centering
%     \includegraphics[width = \linewidth]{images/story_sample.png}
%     \caption{an \auto{} sample. An automatically generated story and corresponding questions and answers about it.}
%     \label{fig:dataset_samples}
% \end{figure*}

%  \begin{figure}
%     \centering
%     \includegraphics[width=\linewidth]{images/nlvr-random.png}
%     \caption{The final scene created by the random choices described in Fig \ref{fig:dataset_samples}.}
%     \label{fig:nlvr}
% \end{figure}

Spatial reasoning is a cognitive process based on the construction of mental representations for spatial objects, relations, and transformations~\cite{clements1992geometry}, which is necessary for many natural language understanding (NLU) tasks such as natural language navigation~\cite{chen2019touchdown, roman-roman-etal-2020-rmm, kim-etal-2020-arramon}, human-machine interaction~\cite{landsiedel2017review, roman-roman-etal-2020-rmm}, dialogue systems \cite{udagawa-etal-2020-linguistic}, and clinical analysis~\cite{datta-roberts-2020-hybrid}. 
% \qn{can you check the emnlp splu workshop and see if you can cite a few more papers?}
% \qn{i suggest that we drop the example below because all it does is to show spatial reasoning is useful, which I think the first sentence is enough}
% For instance, if we tell a robot that ``The milk is in the refrigerator, and the refrigerator is in the kitchen; can you bring me the milk?'', then the robot needs to understand that it should go to the kitchen first.
%
%  Spatial reasoning \st{This understanding}} is necessary for many \qn{\st{downstream} natural language understanding (NLU)} tasks \qn{\st{in natural language understanding and generation like}, such as} natural language navigation~\cite{chen2019touchdown} and human-robot interaction~\cite{landsiedel2017review}. 
% For instance, consider that a machine receives this information ``The milk is in the refrigerator. The refrigerator  is in the kitchen.'' If we ask ``where is the milk'' or ask ``bring the milk,'', the machine should do the spatial reasoning over this knowledge and deduce that the milk is in the kitchen, so, at the first step, it should go to the kitchen. 

Modern language models (LM), e.g., BERT \cite{devlin-etal-2019-bert}, ALBERT \cite{lan2019albert}, and XLNet \cite{yang2019xlnet} have seen great successes in natural language processing (NLP). However, there has been limited investigation into {\em spatial reasoning capabilities of LMs}. To the best of our knowledge, \babi~\cite{weston2015towards}~(Fig \ref{fig:babi}) is the only dataset with direct textual spatial question answering (QA)~(Task 17), but it is synthetic and overly simplified:
(1) The underlying scenes are spatially simple, with only three objects and relations only in four directions.
(2) The stories for these scenes are two short, templated sentences, each describing a single relation between two objects. 
(3) The questions typically require up to two-steps reasoning due to the simplicity of those stories.
% \rk{I added Qiang's text here but it is long.}
%Recently, language models~(LM) like BERT \cite{devlin-etal-2019-bert} have been impressive in solving \qn{NLU (defined above)} tasks like question answering. 
%
%For evaluating LMs' spatial reasoning capabilities over text, to the best of our knowledge, the 20-task bAbI dataset~\cite{weston2015towards} is the only existing dataset that contains direct textual spatial question answering challenge~(task 17). Shortcomings of the bAbI dataset include the simplicity of sentences and questions, the basic syntactic template of describing relations, the lack of describing multiple relations in one sentence, and requiring less than two reasoning steps to find the answers.  Our experiments show that by fine-tuning BERT, this dataset can be solved with $100\%$ accuracy. 

% We provide the first human-generated benchmark, \human{},\footnote{~\em{Spatial Reasoning on Textual Question Answering}} containing four different \rk{spatial reasoning:remove} tasks.

To address these issues, this paper proposes a new dataset, \dataset{}\footnote{\em{SPAtial Reasoning on Textual Question Answering}.}~(see Fig.~\ref{fig:dataset_samples}).
Specifically, 
(1) \dataset{} is built on NLVR's \cite{suhr2017corpus} images containing more objects with richer spatial structures (Fig.~\ref{fig:nlvr}). (2) \dataset{}'s stories are more natural, have more sentences, and richer in spatial relations in each sentence. (3) \dataset{}'s questions require deeper reasoning and have four types: {\em find relation} (FR), {\em find blocks} (FB), {\em choose object} (CO), and {\em yes/no} (YN), which allows for more fine-grained analysis of models' capabilities.

We showed annotators random images from NLVR, and instructed them to describe objects and relationships not exhaustively at the cost of naturalness (Sec.~\ref{sec:human}).
In total, we obtained 1.1k unique QA pair annotations on spatial reasoning, evenly distributed among the aforementioned types.
Similar to \babi{}, we keep this dataset in relatively small scale and suggest to use as little training data as possible.
% \footnote{For instance, the original \babi{} contains 2K QA pairs, although \babi{} is a synthetic dataset and later expanded.} % QN: now I think this can be deleted
% Since transforming from images to textual descriptions inevitably loses information, it is possible that the questions are not answerable based on those stories (although we show later that this happens rarely), we allow for a ``don't know'' (DK) option to these questions.\rk{!}
Experiments show that modern LMs (e.g., BERT) do not perform well in this low-resource setting.

This paper thus proposes a way to obtain distant supervision signals for spatial reasoning (Sec.~\ref{sec:auto}). 
As spatial relationships are rarely mentioned in existing corpora, we take advantage of the fact that spatial language is grounded to the geometry of visual scenes. We are able to automatically generate stories for NLVR images \cite{suhr2017corpus} via our newly designed context free grammars (CFG) and context-sensitive rules.
% Taking advantage of the fact that spatial language is grounded to the geometry of visual scenes, we automatically generate spatially coherent stories of NLVR images \cite{suhr2017corpus}, via our newly designed context free grammars (CFG) and context-sensitive rules.
In the process of story generation, we store the information about all objects and relationships, such that QA pairs can also be generated automatically. In contrast to \babi, we use various spatial rules to infer new relationships in these QA pairs, which requires more complex reasoning capabilities.
Hereafter, we call this automatically-generated dataset \auto{}, and the human-annotated one \human{}.
% Based on the ground truth of NLVR \cite{suhr2017corpus} images, we design a context free grammar (CFG) with additional context sensitive rules to produce coherent stories from randomly selected objects from those images. 
% In the process of story generation, we store the information about all objects and relationships, such that QA pairs can be generated automatically, too. In contrast to \babi, we use various spatial rules to infer new relationships in these QA pairs, which requires more complex reasoning capabilities.
% Hereafter, we call this automatically-generated dataset \auto{}, and the human-annotated one \human{}.

Experiments show that, by further pretraining on \auto{}, we improve LMs' performance on \human{} by a large margin.\footnote{Further pretraining LMs has become a common practice and baseline method for transferring knowledge between tasks \cite{phang2018sentence,zhou2020temporal}. We leave more advanced methods for future work.} The spatially-improved LMs also show stronger performance on two external QA datasets, \babi{} and \boolq{}~\cite{clark2019boolq}: 
BERT further pretrained on \auto{} only requires half of the training data to achieve 99\% accuracy on \babi{} as compared to the original BERT; on boolQ's development set, this model shows better performance than BERT, with 2.3\% relative error reduction.\footnote{To the best of our knowledge, the test set or leaderboard of \boolq{} has not been released yet.}
% given different sizes of training data, BERT shows consistent improvement on \babi{} if further pretrained on \auto{} and in particular, 
% it only requires half of the training data to achieve 99\% accuracy as compared to the original BERT; on \boolq{}, further pretraining BERT on \auto{} shows the best performance on its development set, with 9\% relative error reduction.\footnote{To the best of our knowledge, the test set or leaderboard of \boolq{} has not been released yet.}
%
%
% Our contributions can be summarized as follows. (1) We propose the first human-curated benchmark, \human{}, for spatial reasoning with richer spatial phenomena, in contrast to a prior synthetic dataset \babi{} (Task 17). (2) We design novel CFGs and spatial reasoning rules to obtain indirect supervision data, \auto{}. (3) Further pretraining LMs on \auto{} improves on \human{} by a large margin, and shows state-of-the-art performance on two external QA benchmark datasets, \babi{} and \boolq{}.

\textbf{Our contributions can be summarized as follows.}
First, we propose the first human-curated benchmark, \human{}, for spatial reasoning with richer spatial phenomena than the prior synthetic dataset \babi{} (Task 17). 

Second, we exploit the scene structure of images and design novel CFGs and spatial reasoning rules to automatically generate data (i.e., \auto) to obtain distant supervision signals for spatial reasoning over text.
% \pk{just pointing to the grammar and not pointing to data generation is vague, make ti more concrete }%QN: I don't get your point here; is it still up-to-date? \rk{Yes. what is the problem?} QN: the problem is that i don't understand your problem ... RK: I updated this part. So do you mean what was the problem with the previous one? 
% QN: ok, never mind. the current version looks good to me. i thought your comment was for the current version

Third, \auto{} proves to be a rich source of spatial knowledge that improved the performance of LMs on \human{} as well as on different data domains such as \babi{} and \boolq{}.

\section{Related work}
\label{sec:related}

Question answering is a useful format
% \pk{format: useful format to evaluate the capability ? format looks like to be very low level technical term not actually a learning setting  but I might have missed this usage in the literature}\qn{I remember I had ``format'' here; ``task'' isn't correct here.}
to evaluate machines' capability of reading comprehension~\cite{GBHTM19} and many recent works have been implementing this strategy to test machines' understanding of linguistic formalisms: \citet{HeLeZe15,MSHDZ17,LSCZ17,jia2018tempquestions,ning2020torque,du-cardie-2020-event}. An important advantage of QA is using natural language to annotate natural language, thus having the flexibility to get annotations on complex phenomena such as {\em spatial reasoning}. However, spatial reasoning phenomena have been covered minimally in the existing works.
% \cite{jia2018tempquestions,mishra2018tracking,amini2019mathqa,chen2019codah,tandon2019wiqa}.
% mishra2018tracking, amini2019mathqa, chen2019codah are not suitable

% use QA as a format to probe systems' understanding of certain linguistic formalisms~\cite{jia2018tempquestions,mishra2018tracking,amini2019mathqa,chen2019codah,tandon2019wiqa}.
% \qn{cite QASRL, QAMR, QA-RE, McTACO, TORQUE, Clardie's event extraction via QA in emnlp2020}. 
% An important advantage of it is that it uses natural language to annotate natural language, thus having the flexibility to get annotations on complex phenomena such as {\em spatial reasoning}.
% However, spatial reasoning phenomena have been covered minimally in these existing works.

To the best of our knowledge, Task 17 of the \babi{} project~\cite{weston2015towards} is the only QA dataset focused on textual spatial reasoning (examples in Appendix~\ref{sec:babiboolq}). However, \babi{} is synthetic and does not reflect the complexity of the spatial reasoning in natural language. Solving Task 17 of \babi{} typically does not require sophisticated reasoning, which is an important capability emphasized by more recent works (e.g., \citet{dua2019drop,MultiRC2018,yang2018hotpotqa,DLMSG19,ning2020torque}).

Spatial reasoning is arguably more prominent in multi-modal QA benchmarks, e.g., NLVR~\cite{suhr2017corpus}, VQA~\cite{antol2015vqa}, GQA~\cite{hudson2019gqa}, CLEVR~\cite{johnson2017clevr}. However, those spatial reasoning phenomena are mostly expressed naturally through images, while this paper focuses on studying spatial reasoning on natural language. 
Some other works on visual-spatial reasoning are based on geographical information inside maps and diagrams~\cite{huang2019geosqa} and navigational instructions~\cite{chen2019touchdown,anderson2018vision}.

As another approach to evaluate spatial reasoning capabilities of models, a dataset proposed in \citet{ghanimifard2017learning} generates a synthetic training set of spatial sentences and evaluates the models' ability to generate spatial facts and sentences containing composition and decomposition of relations on grounded objects. 

% \qn{below may be unnecessary}
% Training LMs on specific tasks and evaluating them based on their performance before and after that is also investigated in recent research on temporal reasoning~\cite{ning2020torque}, common-sense reasoning~\cite{rajani2019explain}, and mathematical reasoning~\cite{rabe2020mathematical}. However, previous research mainly focused on collecting human-annotated data, which is very expensive, and we propose a novel approach to create a supervision task based on an automatic process.

% As another approach to evaluate spatial reasoning capabilities of models, this dataset proposed in \cite{ghanimifard2017learning} generates a synthetic training set of spatial sentences and evaluates the models' ability to generate spatial facts and sentences containing composition and decomposition of relations on grounded objects. 

%!TEX root=naacl2021.tex
\section{\human}
\label{sec:human}
To mitigate the aforementioned problems of Task 17 of \babi{}, i.e., simple scenes, stories, and questions, we describe the data annotation process of \human{}, and explain how those problems were addressed in this section.

First, we randomly selected a subset of NLVR images, each of which has three blocks containing multiple objects (see Fig~\ref{fig:nlvr}). The scenes shown by these images are more complicated than those described by \babi{} because (1) there are more objects in NLVR images; (2) the spatial relationships in NLVR are not limited to just four relative directions as objects are placed arbitrarily within blocks.

% how stories are created
Second, two student volunteers produced textual description of those objects and their corresponding spatial relationships based on these images. 
Since the blocks are always horizontally aligned in each NLVR image, to allow for more flexibility, annotators could also rearrange these blocks (see Fig.~\ref{fig:story}).
Relationships between objects within the same block can take the forms of relative direction~(e.g., left or above), qualitative distance~(e.g., near or far), and topological relationship~(e.g., touching or containing).

However, we instructed the annotators not to describe all objects and relationships, (1) to avoid unnecessarily verbose stories, and (2) to intentionally miss some information to enable more complex reasoning later. Therefore, 
annotators describe only a random subset of blocks, objects, and relationships.
% \qn{it's not necessary, but if you have time, get some statistics on average number of tokens in your stories, number of objects used, and number of relations reported.}\rk{this is not applicable here. If we want to talk about this we should try it in the auto data section}

% how questions and answers are created
To query more interesting phenomena, annotators were then encouraged to write questions requiring detecting relations and reasoning over them using multiple spatial rules.
A spatial rule can be one of the transitivity~($A \rightarrow B, B \rightarrow C \Rightarrow A \rightarrow C$), symmetry~($A \rightarrow B \Rightarrow B \rightarrow A$), converse~($(A,\ R,\ B)\Rightarrow (B,\ reverse(R),\ A)$), inclusion~($obj1\ in\ A$), and exclusion~($obj1\ not\ in\ B$) rules.

There are four types of questions~(\qtype). (1) {\em FR}: find relation between two objects. (2) {\em FB}: find the block that contains certain object(s). (3) {\em CO}: choose between two objects mentioned in the question that meets certain criteria. (4) {\em YN}: a yes/no question that tests if a claim on spatial relationship holds.
% All four types of questions are formulated as multiple-choice questions.
% Specifically, 
\begin{figure}[t]
    \centering
    \includegraphics[width=0.5\linewidth]{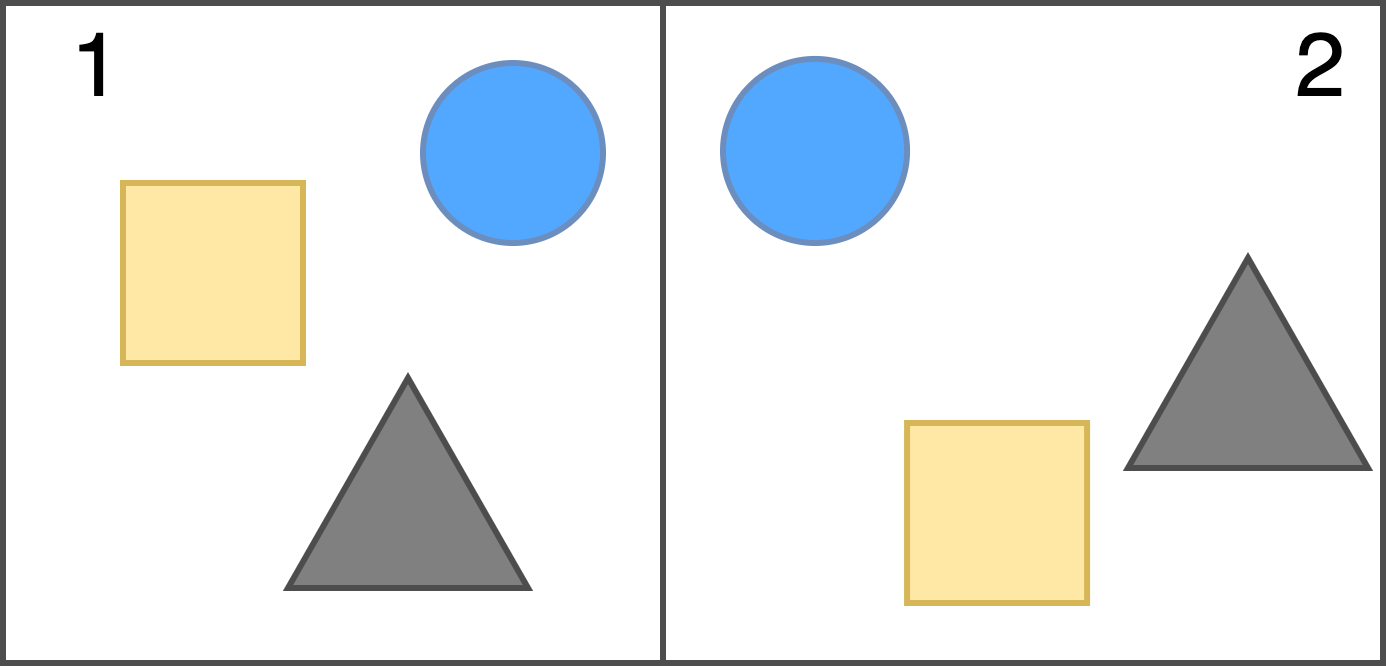}
    \caption{For ``A blue circle is above a big triangle. To the left of the big triangle, there is a square,'' if the question is: ``Is the square to the left of the blue circle?'', the answer is neither Yes nor No. Thus, the correct answer is ``Do not Know'' (DK)  in our setting.}
    \label{fig:DK}
\end{figure}

FB, FR, and CO questions are formulated as multiple-choice questions\footnote{CO can be considered as both single-choice and multiple-choices question.} and receive a list of candidate answers, and YN questions' answer is choosing from Yes, No, or ``DK''~(Do not Know). The ``DK'' option is due to the open-world assumption of the stories, where if something is not described in the text, it is not considered as false~(See Fig. \ref{fig:DK}).

% \rk{Please see Table \ref{tab:templates} in the Appendix for more details: this table contains the templates for automatic generated questions,  It shouldn't be here.}.

% The FR~(Find Relation) questions ask about the relationships between two objects. The FB~(Find block) questions objective is to find blocks that contain an object (or a group of objects) with specific features. The CO~(Choose object) questions provide two objects and ask which one has a relation to another object with particular features.  The last question type is YN, a Yes/No question that  asks whether the relation between two objects holds. 

% Finally, the correct answers are one or a set of options in a multiple-choice list. The FB, FR, and CO question types receive a list of all candidate answers~(see table \ref{tab:templates} in Appendix), and the YN question type can only have one correct answer Yes, No, or ``DK''~(Do not Know). Same as SQuADv2~\cite{rajpurkar2018know}, we use unanswerable questions that their answers may not be provided in the text, and a reliable model should answer ``None'' or ``DK'' when necessary. The ``DK'' answer is due to the open-world assumption of the stories, where if something is not described in the text, it is not considered false.~(See fig \ref{fig:DK})

% special notes

% \qn{I find this a minor detail; can we remove it?} The annotators also are asked to add the spatial rules that are needed for answering each question.

% statistics
Finally, annotators were able to create 1.1k QA pairs on spatial reasoning on the generated descriptions, distributed among the aforementioned types. 
% \rk{I removed "and roughly evenly"} 
We intentionally keep this data in a relatively small scale due to two reasons. First, there has been some consensus in our community that modern systems, given their sufficiently large model capacities, can easily find shortcuts and overfit a dataset if provided with a large training data \cite{gardner2020evaluating,sen-saffari-2020-models}.
Second, collecting spatial reasoning QAs is very costly: The two annotators spent 45-60 mins on average to create a single story with 8-16 QA pairs. 
We estimate that \human{} costed about 100 human hours in total.
% They spend 45-60 minutes~(on average) on creating each story and 8-16 related questions and answers.
% who spend 45-60 minutes~(on average) creating a story and 8-16 related questions and answers.
The expert performance on 100 examples of \human{}'s test set measured by their accuracy of answering the questions is 92\% across four \qtype{}s on average, indicating its high quality.

\begin{table}
    \centering
    \resizebox{\linewidth}{!}{%
    \begin{tabular}{|l|cccc|c|} 
        \hline
        Sets        & FB    & FR    & YN    & CO &Total    \\ 
        \hline
        \human:&&&&&\\
        \quad Test & 104  & 105  & 194  & 107 &510    \\
        \quad Train & 154  & 149  & 162  & 151   &616  \\\hline
        \auto:&&&&& \\
        \quad Seen Test & 3872  & 3712  & 3896  & 3594  & 15074   \\
        \quad Unseen Test & 3872  & 3721  & 3896  & 3598 &15087    \\
        \quad Dev         & 3842  & 3742  & 3860  & 3579  &15023   \\
        % Train       & 7803  & 7315  & 7501  & 6520     \\
        \quad Train    & 23654 & 23302 & 23968 & 22794 & 93673  \\
        \hline
    \end{tabular}
    }
    \caption{Number of questions per \qtype}
    \label{tab:num_question}
\end{table}

%!TEX root=naacl2021.tex
\section{Distant Supervision: \auto}
\label{sec:auto}
% overview and challenge
Since human annotations are costly, it is important to investigate ways to generate distant supervision signals for spatial reasoning. 
However, unlike conventional distant supervision approaches (e.g., \citet{mintz2009distant,zeng2015distant,zhou2020temporal}) where distant supervision data can be selected from large corpora by implementing specialized filtering rules, spatial reasoning does not appear often in existing corpora. Therefore, similar to \human{}, we take advantage of the ground truth of NLVR images, design CFGs to generate stories, and use spatial reasoning rules to ask and answer spatial reasoning questions. This automatically generated data is called \auto{}, and below we describe its generation process in detail.

% story
\paragraph{Story generation} Since NLVR comes with structured descriptions of the ground truth locations of those objects, we were able to choose random blocks and objects from each image programmatically. The benefit is two-fold.
First, a random selection of blocks and objects allows us to create multiple stories for each image; second, this randomness also creates spatial reasoning opportunities with missing information.

Once we decide on a set of blocks and objects to be included, we determine their relationships: Those relationships between blocks are generated randomly; as for those between objects, we refer to the ground truth of these images to determine them.

\begin{figure}[h]
	\centering
	\begin{subfigure}[b]{\linewidth}
		\includegraphics[width=\linewidth]{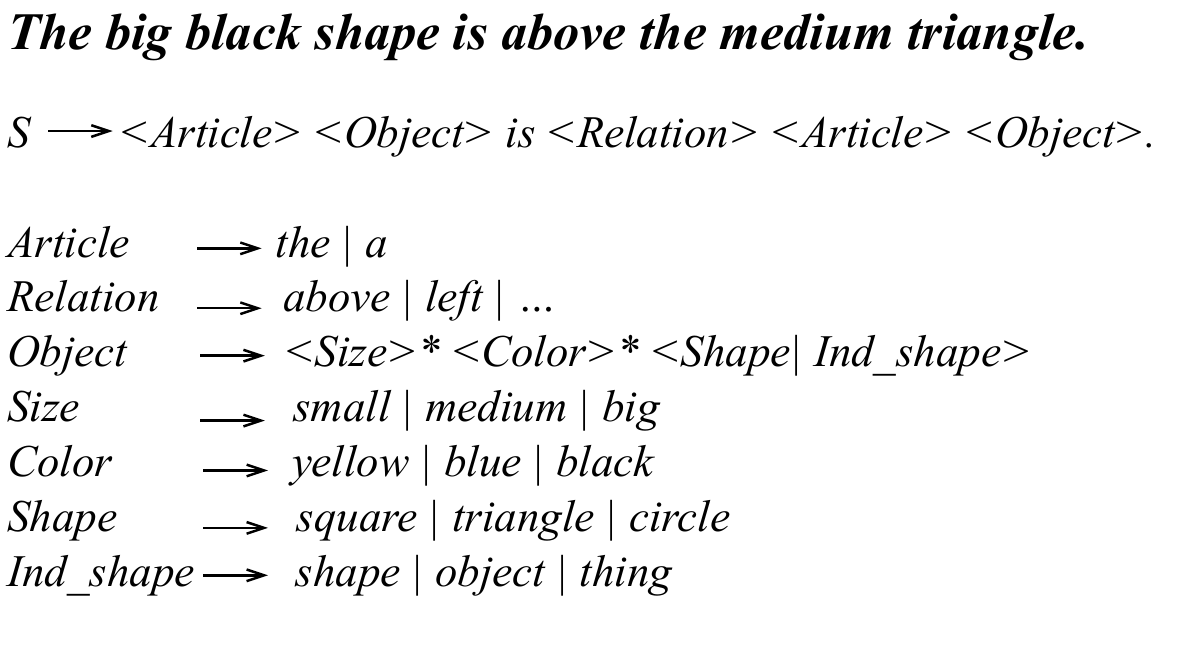}
		\caption{Part of the grammar describing relations between objects
% 		A sample of grammar to describe relation between objects. 
% 		\qn{do you want to refer to this type of relations as ``relative relations'', in contrast to your ``nested relations'' below?}\rk{relative relations is the name of that four LEFT. RIGHT, ... relations. So we cannot do this. I think just relation is fine.} \qn{so your fig a is ``relations'' and fig b is ``nested relations''?}\rk{yes the first one is a sample of describing a relation between two objects.}
        }
		\label{fig:cfg1}
	\end{subfigure}
%	\hfill
	\begin{subfigure}[b]{\linewidth}%{0.8\linewidth}
			\includegraphics[width=\linewidth]{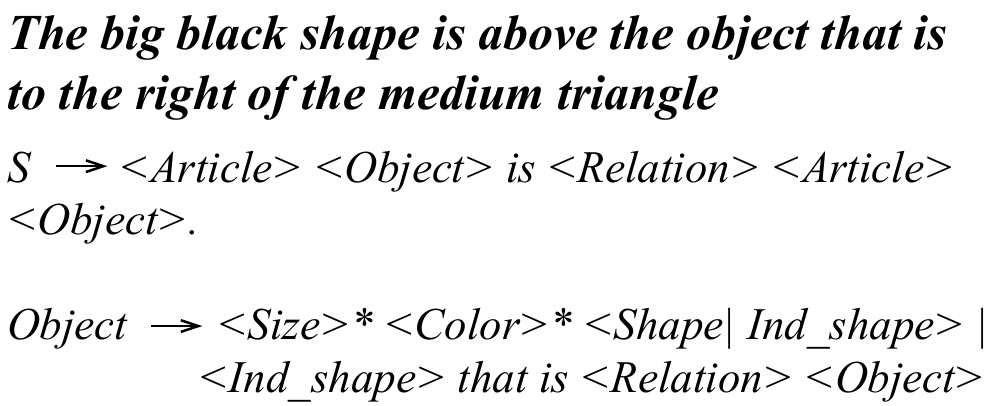}
			\caption{Part of the grammar describing nested relationships.
% 			A sample of grammar to describe nested relations for an object.
			}
			\label{fig:cfg2}
    \end{subfigure}
 	\caption{Two parts of our designed CFG}
	\label{fig:cfg}
\end{figure}

Now we have a scene containing a set of blocks and objects and their associated relationships. To produce a story for this scene, we design CFGs to produce natural language sentences that describe those blocks/objects/relationships in various expressions~(see Fig.~\ref{fig:cfg} for two portions of our CFG describing relative and nested relations between objects).

Being grounded to visual scenes guarantees spatial coherency in a story, and using CFGs helps to have correct sentences (grammatically) and various expressions. We also design context-sensitive rules to limited options for each CFG's variable based on the chosen entities (e.g. black circle), or what is described in the previous sentences (e.g. Block A has \textit{a} circle. \textit{The} circle is below \textit{a} triangle.)

\begin{figure}
    \centering
    \includegraphics[width=0.9\linewidth, angle=0]{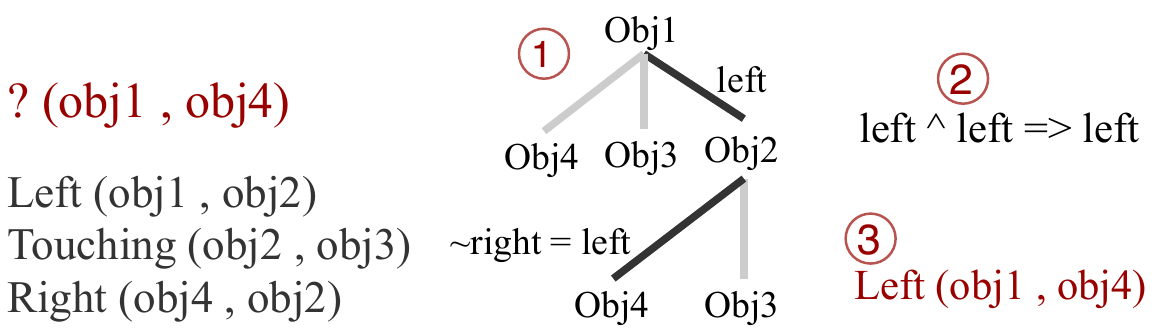}
    \caption{Find the implicit relation between $obj1$ and $obj4$ by {\em Transitivity} rule. (1) Find a set of objects that have a relation with $obj1$. Continue the same process on the new set until $obj4$ is found. (2) Get the union of the intermediate relations between these two objects and it is the final answer.}
    \label{fig:transitivity_rel}
\end{figure}

% question
\paragraph{Question generation} To generate questions based on a passage, there are rule-based systems~\cite{heilman2009question,labutov2015deep}, neural networks~\cite{du2017learning}, and their combinations.
% of these two methods by using rules along with semantic role labeling~(SRL)~
\cite{dhole2020syn}. However, in our approach, during generating each story, the program stores the information about the entities and their relationships. 
Thus, without processing the raw text, which is error-prone, we generate questions by only looking at the stored data.
% Thus, without any need to process the raw text, which can make errors, we generate questions by only looking at the stored data.
%

The question generation operates based on four primary functionalities, \textit{Choose-objects}, \textit{Describe-objects}, \textit{Find-all-relations}, and \textit{Find-similar-objects}. These modules are responsible to control the logical consistency, correctness, and the number of steps required for reasoning in each question.

\textit{\textbf{Choose-objects}} 
randomly chooses up to three objects from the set of possible objects in a story under a set of constraints such as preventing selection of similar objects, or excluding objects with relations that are directly mentioned in the text.
% randomly chooses one, two, or three objects by applying special rules to prevent choosing similar objects or objects with direct relations (to have multi-step reasoning). 

\textit{\textbf{Describe-Objects}}
generates a mention phrase for an object using parts of its full name (presented in the story). The generated phrase is either pointing to a unique object or a group of objects such as "the big circle," or "big circles." To describe a unique object, it chooses an attribute or a group of attributes that apply to a unique object among others in the story. To increase the steps of reasoning, the description may include the relationship of the object to other objects instead of using a direct unique description. For example, "the circle which is above the black triangle."

\textit{\textbf{Find-all-relations}}
completes the relationship graph between objects by applying a set of spatial rules such as transitivity, symmetry, converse, inclusion, and exclusion on top of the direct relations described in the story.
As shown in Fig.~\ref{fig:transitivity_rel}, it does an exhaustive search over all combinations of the relations that link two objects to each other.

% For instance, to find the relationship of ``Big Circle'' and ``Triangle'' in the story "The big circle is in block B. Block B is above block A. Block A has a triangle,", the algorithm should link relations from \textit{``(big circle, in (NTPP), block B), (block B, above (ABOVE), block A), (block A, has (TPP), triangle)''} and find \textit{``above(ABOVE)''} as the relationship between the objects of interest.

% uses spatial rules  such as transitivity, symmetry, converse, inclusion, and conclusion. As shown in Fig.~\ref{fig:transitivity_rel}, computing indirect relations based on the story requires considering all relationships of two objects with an arbitrary number of intermediate ones.

\textit{\textbf{Find-similar-objects}}
finds all the mentions matching a description from the question to objects in the story. For instance, for the question "is there any blue circle above the big blue triangle?", this module finds all the mentions in the story matching the description ``a blue circle''.
% Since we are dealing with the natural language questions, a phrase may mention more than one objects (e.g., having "the big circle, the small yellow circle" and the mention was "the circle"), we find all possible mentioned objects and do the reasoning over all of them.

% \pk{You use the above four modules to generate four Q-types?\rk{to generate questions in general} Then you describe how you use those modules for generating question in the following paragraph? this was no very clear, maybe revise this part again and make this clear.}\rk{It is two different paragraph. The first one talking about the different qtypes and the second one start talking about the main Question generation. I changed the place of qtype description and added it below. so now at first I described four modules and then talked about how we used these modules in different qtypes.}

Similar to the \human{}, we provide four \qtype{}s FR, FB, CO, and YN. 
To generate FR questions, we choose two objects using \textit{Choose-objects} module and question their relationships. The YN \qtype{} is similar to FR, but the question specifies one relationship of interest chosen from all relation extracted by \textit{Find-all-relations} module to be questioned about the objects. Since most of the time, Yes/No questions are simpler problems, we make this question type more complex by adding quantifiers (adding ``all'' and ``any''). These quantifiers help to evaluates the models' capability to aggregate relations between more than two objects in the story and do the reasoning over all find relations to find the final answer. In FB \qtype{}, we mention an object by its indirect relation to another object using the nested relation in \textit{Describe-objects} module and ask to find the blocks containing or not containing this object. 
Finally, the CO question selects an anchor object (\textit{Choose-objects}) and specifies a relationship ( using \textit{Find-all-relations}) in the question. Two other objects are chosen as candidates to check whether the specified relationship holds between them and the anchor object. We tend to force the algorithm to choose objects as candidates that at least have one relationship to the anchor object.
 To see more details about different question' templates see Table \ref{tab:templates} in the Appendix.
 
% \pk{You did not explicitly refer back to the above-mentioned modules to make it clear how those are used in generating the types of questions.}

% answer

\paragraph{Answer generation} We compute all direct and indirect relationships between objects using \textit{Find-all-relations} function and based on the \qtype s generate the final answer.  

For instance, in YN \qtype{} if the asked relation exists in the found relations, the answer is "Yes", if the inverse relation exists it must be "No", and otherwise, it is "DK"\footnote{The \auto{} generation code and the file of dataset are available at \url{https://github.com/HLR/SpartQA_generation}}.

\subsection{Corpus Statistics}
We generate the train, dev, and test set splits based on the same splits  of the images in the NLVR dataset. %\rk{I want to mention that we generate train set base on the images in the train set. How do I mention thaa?}
On average, each story contains 9 sentences (Min:3, Max: 22) and 118 tokens (Min: 66, Max: 274). Also, the average tokens of each question (on all \qtype~) is 23 (Min:6, Max: 57). 

Table \ref{tab:num_question} shows the total number of each question type in \auto{}~(Check Appendix to see more statistic information about the labels in Tab \ref{tab:number_of_choices}.)

% \qn{The generation of \auto{} have followed that of \human{} closely and is also based on images from NLVR. Note the generation method described in this section is generally applicable to auto generation of QA data on spatial reasoning}

%!TEX root=naacl2021.tex

\section{Models for Spatial Reasoning over Language}
\label{Sec:Architecture}
This section describes the model architectures on different \qtype{}s: FR, YN, FB, and CO. All \qtype{}s can be cast into a sequence classification task, and the three transformer-based LMs tested in this paper, BERT \cite{devlin-etal-2019-bert}, ALBERT \cite{lan2019albert}, and XLNet \cite{yang2019xlnet}, can all handle this type of tasks by classifying the representation of [CLS], a special token prepended to each target sequence (see Appendix \ref{sec:lm-arch}). Depending on the \qtype, the input sequence and how we do inference may be different.

FR and YN both have a predefined label set as candidate answers, and their input sequences are both the concatenation of a story and a question. While the answer to a YN question is a single label chosen from {\em Yes, No}, and {\em DK}, FR questions can have multiple correct answers. Therefore, we treat each candidate answer to FR as an independent binary classification problem, and take the union as the final answer.
As for YN, we choose the label with the highest confidence (Fig \ref{fig:modelyn}).

As the candidate answers to FB and CO are not fixed and depend on each story and its question the input sequences to these \qtype{}s are concatenated with each candidate answer. Since the defined YN and FR model has moderately less accurate results on FB and CO \qtype s, we add a LSTM~\cite{hochreiter1997long} layer to improve it.
% we modify the described model to contain a candidate answer as input. 
Hence, to find the final answer, we run the model with each candidate answer and then apply an LSTM layer on top of all token representations. Then, we use the last vector of the LSTM outputs for classification (Fig \ref{fig:modelfb}). 
% \begin{figure}
%     \centering
%     \includegraphics{images/CO.png}
%     \caption{Langugae model fine-tuning on FB and CO tasks}
%     \label{fig:CO}
% \end{figure}
The final answers are selected based on Eq.~\eqref{Formula:Answer2}. 

\begin{equation}
\label{Formula:Answer2}
    \begin{split}
        x_{i} &= [s, c_{i}, q] \\
        \vec{T_i} &= [\vec{t^{i}_{1}}, ..., \vec{t^{i}_{m^{i}}}] = LM(x_{i}) \\
        [{\vec{h}^{i}_1}, ..., {\vec{h}^{i}_{m^i}}] &= \textrm{LSTM}(\vec{T_i}) \\
        \vec{y_{i}} = [y^{0}_{i}, y^{1}_{i}] &= \textrm{Softmax}(\vec{h}^{i^T}_{m^i}W)) \\
        \text{Answer} &= \{c_{i}| \argmax_j(y^j_{i}) = 1\} 
    \end{split}
\end{equation}
where $s$ is the story, $c_{i}$ is the candidate answer, $q$ is the question, $[\ ]$ indicates the concatenation of the listed vectors, and $m_{i}$ is tokens' number in $x_{i}$. The parameter vector, $W$, is shared for all candidates.

\subsection{Training and Inference}
\label{Sec:Inference}
We train the models based on the summation of the cross-entropy losses of all binary classifiers in the architecture. For FR and YN \qtype{}s, there are multiple classifiers, while there is only one classifier used for CO and FB \qtype{}s.

We remove inconsistent answers in 
% a 
post-processing 
% step 
for FR and YN \qtype{}s during inference phase. % at prediction time. 
For instance on FR, {\em left} and {\em right} relations between two objects cannot be valid at the same time. For YN, as there is only one valid answer amongst the three candidates, we select the candidate with the maximal 
% maximum 
predicted probability of being the true answer.

%!TEX root=naacl2021.tex
\section{Experiments}
\label{sec:Experiments}
% \subsection{Setup}
% \label{Sec:Setup}

\begin{table*}[t]
    \centering
      \begin{tabular}{|l|l|c|c|c|c|c|}
        \hline
        \#&Model  & FB  & FR & CO & YN & Avg\\ %\cline{3-14} 
        \hline
        1&Majority %baseline       
             & 28.84 &24.52 & 40.18 &53.60 & 36.64\\
        \hline
        %  \multirow{3}{*}{EXP 1}
        %   &Seen test    & 27.35 &37.73 & 40.38 &67.85 & 43.32\\ \cline{2-2}
        %   &Unseen test   & 20.5  &32.64 & 39.71 &67.74 & 40.14\\ \cline{2-2}
            2&BERT%$_{QA}$        
             & 16.34 &20 & 26.16 &45.36 & 30.17\\
        %  EXP 2 &Human& 87.13&69.38& 44.24&85.68\\\hline
        %   \multirow{3}{*}{EXP 2}
        %   &Seen test    & 26.79 &34.13 & 41.33 &67.91 & 42.54\\ \cline{2-2}
        %   &Unseen test   & 23.2  &29.94 & 40.72 &65.82 & 39.92\\ \cline{2-2}
          3&BERT (Stories only; MLM) %$_{MLM}$--{\small \auto{}(T)}
            & 21.15 &16.19   & 27.1 &\textbf{51.54} & 32.90\\
          
        4&BERT (\auto{}; MLM) %$_{MLM}$--{\small \auto{}(T+Q)}         
        & 19.23 &29.54   & \textbf{32.71} &47.42 & 34.88\\
        %   \multirow{3}{*}{EXP 3}
        %   &Seen test    & \textbf{86.85} &\textbf{85.86} & \textbf{71.47} &\textbf{78.29} & \textbf{80.61}\\ \cline{2-2}
        %   &Unseen test   & \textbf{69.77} &\textbf{74.61} & \textbf{61.18} &\textbf{78.06} & \textbf{70.9}\\ \cline{2-2}
          5&BERT (\auto{})%$_{QA}$ + \small \auto
          & \textbf{62.5} &\textbf{46.66} & \textbf{32.71} &47.42  & \textbf{47.25}\\\hline
          
          6&Human         & 91.66 &95.23 & 91.66 &90.69  &92.31\\\hline
      \end{tabular}
    \caption{
    \textbf{Further pretraining BERT on \auto{} improves accuracies on \human{}}. All systems are fine-tuned on the training data of \human{}, but Systems~3-5 are also further pretrained in different ways. System 3: further pretrained on the stories from \auto{} as a masked language model (MLM) task. System 4: further pretrained on both stories and QA annotations as MLM. System 5: the proposed model that is further pretrained on \auto{} as a QA task. Avg: The micro-average on all four \qtype{}s.
    % Evaluating BERT on \human{}  by four experiments. 1) BERT$_{QA}$, 2) BERT$_{MLM}$--{\small \auto{}(T)}: fine-tuned BERT on \auto{}'s text as MLM task, 3) BERT$_{MLM}$--{\auto{}(T+Q)}: fine-tuned BERT on \auto{}'s text+questions as MLM task, and 4)BERT$_{QA}$--{\small \auto{}}:  fine-tuned BERT on \auto{} as QA task. All models are fine-tuned on \human{} training samples. Avg*: This column shows the weighted average result on all four tasks. All numbers are the accuracy of the models' performance.
    }
    \label{tab:experiments}
\end{table*} 

As fine-tuning LMs has become a common baseline approach to knowledge transfer from a source dataset to a target task, including but not limited to \citet{phang2018sentence,zhou2020temporal,he2020foreshadowing}, we study the capability of spatial reasoning of modern LMs, specifically BERT, ALBERT, and XLNet, after fine-tuning them on \auto{}. This fine-tuning process is also known as {\em further pretraining}, to distinguish with the fine-tuning process on one's target task. 
% remove if we don't have space
It is an open problem to find out better transfer learning techniques than simple further pretraining, as suggested in \citet{he2020quase,2020unifiedqa}, which is beyond the scope of this work.
All experiments use the models proposed in Sec.~\ref{Sec:Architecture}. We use AdamW~\cite{loshchilov2017decoupled} with $2\times 10^{-6}$ learning rate and Focal Loss~\cite{lin2017focal} with $\gamma = 2$ for training all the models.\footnote{All codes are available at \url{https://github.com/HLR/SpartQA-baselines}}
% \qn{is this correct?}\rk{yes}

\subsection{Further pretraining on \auto{} improves spatial reasoning}

Table~\ref{tab:experiments} shows performance on \human{} in a low-resource setting, where 0.6k QA pairs from \human{} are used for fine-tuning these LMs and 0.5k for testing (see Table~\ref{tab:num_question} for information on this split).\footnote{Note this low-resource setting can also be viewed as a spatial reasoning probe to these LMs \cite{tenney2019you}.}
During our annotation, we found that the description of ``near to '' and ``far from'' varies largely between annotators. Therefore, we ignore these two relations from FR \qtype{} in our evaluations.
% during evaluating \human{} examples to be consistent.

In Table~\ref{tab:experiments}, System~5, BERT (\auto), is the proposed method of further pretraining BERT on \auto{}. We can see that System~2, the original BERT, performs consistently lower than System~5, indicating that having \auto{} as a further pretraining task improves BERT's spatial understanding.

\begin{table}[h]
    \centering
    \begin{tabular}{|l|c|}
        \hline
        Model & $F_1$ \\ \hline
         Majority& 35 \\ \hline
         BERT& 50 \\
         BERT (Stories only; MLM) & 53\\
         BERT (\auto{}; MLM)& 48\\
         BERT (\auto{}) & 48\\
         \hline
    \end{tabular}
    \caption{Switching from accuracy in Table~\ref{tab:experiments} to $F_1$ shows that the models are all performing better than the majority baseline on YN \qtype. 
    % Evaluating models on YN \qtype{} with F1-measure.
    }
    \label{tab:f1-YN}
\end{table}

\begin{table*}[t]
    \centering
    \resizebox{\textwidth}{!}{
    \begin{tabular}{|c|l|ccc|ccc|ccc|ccc|}
        \hline
        \multirow{2}{*}{\#}&
        \multirow{2}{*}{Models} & \multicolumn{3}{c}{FB}  & \multicolumn{3}{c}{FR} & \multicolumn{3}{c}{CO} & \multicolumn{3}{c|}{YN}\\ \cline{3-14}
         &&Seen& Unseen& Human*&Seen& Unseen& Human*&Seen& Unseen& Human*&Seen& Unseen& Human*\\ \hline
        %   &&&&&&&&&&&&&\\
          1&Majority &48.70&48.70 & 28.84& 40.81& 40.81 & 24.52& 20.59 &20.38 & 40.18 &49.94 &49.91& \textbf{53.60}\\ 
          \hline
        %   2&BERT_{QA}&  & 13.51& 16.34&&& 20&&& 26.16&&& 45.36&30.17\\
        %   &&&&&&&&&&&&&&\\
        %   BERT + auto-spartQA & - & -& 21.15&-&-& 16.19&-&-& 27.1&-&-& 51.54\\
        %   \multicolumn{1}{|r|}{text_{MLM}}&&&&&&&&&&&&\\
        %   3&BERT_{MLM}&  && &&& 16.19&&& 27.1&&& 51.54&32.90\\
        %   &\small \auto{}(T) &&&&&&&&&&&&&\\
        %   4&BERT_{MLM}&  & & 19.23&&& 29.54&&& 32.71&&& 47.42&34.88\\
        %   &\small \auto{}(T+Q) &&&&&&&&&&&&&\\
        %   BERT + auto-spartQA & - & -& 19.23&-&-& 29.54&-&-& 32.71&-&-& 47.42\\
        %   \multicolumn{1}{|r|}{text+question_{MLM}}&&&&&&&&&&&&\\
        %   BERT^{S}_{QA}& - & -& 62.5&-&-& 46.66&-&-& 32.71&-&-& 47.42\\
        %   &&&&&&&&&&&&\\
        %   BERT +  & - & -& 62.5&-&-& 46.66&-&-& 32.71&-&-& 47.42\\
        %   \multicolumn{1}{|r|}{auto-spartQA_{(QA)}}&&&&&&&&&&&&\\

         2&BERT & 87.13 & 69.38& 62.5&85.68&73.71& 46.66&71.44&61.09& 32.71&78.29&76.81& 47.42\\ 
        %   &\small \auto{}&&&&&&&&&&&&\\
         3&ALBERT &97.66&83.53& 56.73&91.61 &83.70& 44.76 & 95.20 & 84.55 & 49.53& 79.38&75.05& 41.75\\  
        %  \multicolumn{1}{|r|}{auto-spartQA_{(QA)}}
        %  &\small \auto{}&&&&&&&&&&&&\\
          4&XLNet &\textbf{98.00}&\textbf{84.85} & \textbf{73.07}& \textbf{94.60} &\textbf{91.63}& \textbf{57.14}& \textbf{97.11} &\textbf{90.88} & \textbf{50.46}&\textbf{79.91} &\textbf{78.54}& 39.69\\
        %   \multicolumn{1}{|r|}{auto-spartQA_{(QA)}}
        %   &\small \auto{}&&&&&&&&&&&&\\
          \hline
        %  \multirow{2}{*}{RoBERTa}& 8k& & & & & & & & &\\ \cline{2-2}
        %   &24k & & & & & & & & &\\ \hline
        %  \multirow{2}{*}{LSTM}& 8k& & & & & & & & &\\ \cline{2-2}
        %   &24k & & & & & & & & &\\ \hline
            % &&&&&&&&&&&&\\
          5&Human & \multicolumn{2}{c}{85}&91.66&\multicolumn{2}{c}{90}&95.23&\multicolumn{2}{c}{94.44}&91.66&\multicolumn{2}{c}{90}&90.69\\ \hline
    \end{tabular}
    }
    \caption{\textbf{Spatial reasoning is challenging}. We further pretrain three transformer-based LMs, BERT, ALBERT, and XLNet, on \auto{}, and test their accuracy in three ways: {\em Seen} and {\em Unseen} are both from \auto{}, where {\em Unseen} has applied minor modifications to its vocabulary; to get those {\em Human} columns, all models are fine-tuned on \human{}'s training data.
    % Fine-tuning three baseline LMs (BERT, ALBERT, and XLNET) on \auto, and test on three benchmarks, Seen, Unseen and Human~(\human) sets. \textbf{Human*}: For Human result all models are fine-tuned on \human{} training samples. 
    Human performance %The result of Human 
    on {\em Seen} and {\em Unseen} is the same since the changes applied to {\em Unseen} does not affect human reasoning.}
    \label{tab:evaluation}
\end{table*}

\begin{figure}[h]
    \centering
    \includegraphics[width = 0.9\linewidth]{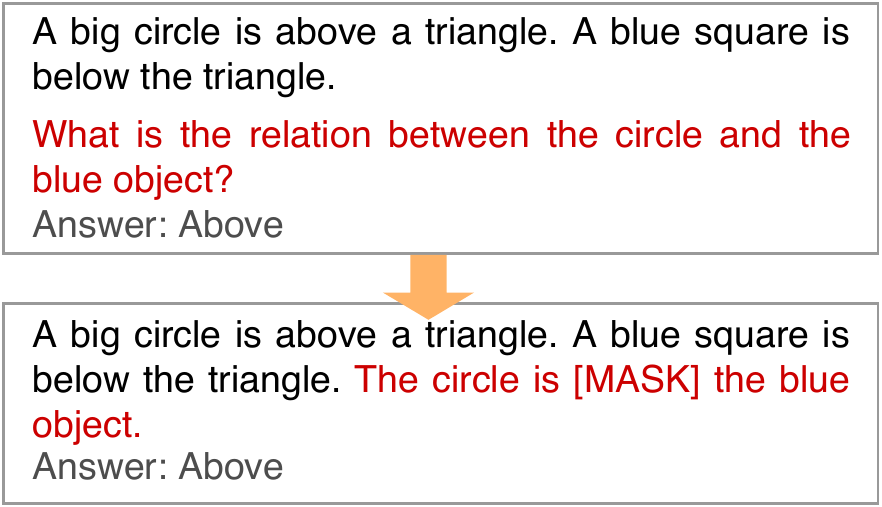}
    \caption{Convert a triplet of (paragraph, question, answer) into a single piece of text for the MLM task. }
    \label{fig:masking}
\end{figure}

In addition, we implement another two baselines. %\rk{compare what? System~5?} to \rk{another: other?} two baselines. 
System~3, BERT (Stories only; MLM): further pretraining BERT only on the stories of \auto{} as a masked language model (MLM) task; System~4, BERT (\auto{}; MLM): we convert the QA pairs in \auto{} into textual statements and further pretrain BERT on the text as an MLM (see Fig.~\ref{fig:masking} for an example conversion). 
% \rk{

To convert each question and its answer into a sentence, we utilize static templates for each question type which removes the question words and rearranges other parts into a sentence.
% }
% To convert questions into the sentence format, we changed the format of questions template~(see Table~\ref{tab:templates}) to a sentence and 
% \qn{it's better to have some short descriptions on how this conversion is done automatically}

We can see that System~3 slightly improves over System~2, an observation consistent with many prior works that seeing more text generally helps an LM (e.g., \citet{gururangan2020dont}). The significant gap between System~3 and the proposed System~5 indicates that supervision signals come more from our annotations in \auto{} rather than from seeing more unannotated text.
System~4 is another way to make use of the annotations in \auto{}, but it is shown to be not as effective as further pretraining BERT on \auto{} as a QA task.

While the proposed System~5 overall 
performs better than the other three baseline systems, one exception is its accuracy on YN, which is lower than that of System~3. 
Since all systems' YN accuracies are also lower than the majority baseline\footnote{which predicts the label that is most common in each set of \dataset{}}, we hypothesize that this is due to imbalanced data. To verify it, we compute the $F_1$ score for YN \qtype{} in Table~\ref{tab:f1-YN}, where we see all systems effectively achieve better scores than the majority baseline. However, further pretraining BERT on \auto{} still does not beat other baseline systems, which implies that straightforward pretraining is not necessarily helpful in capturing the complex reasoning phenomena required by YN questions.

The human performance is evaluated on 100 random questions from each \auto{} and \human{} test set. The respondents are graduate students that were trained by some examples of the dataset before answering the final questions. We can see from Table~\ref{tab:experiments} that all systems' performances fall behind human performance by a large margin. We expand on the difficulty of \textsc{SpartQA} in the next subsection.

\begin{table*}[t]
    \centering
    \resizebox{\textwidth}{!}{%
    \begin{tabular}{|l|c|cc|cc|cc|}
        \hline
        \multirow{2}{*}{Models} &  \multicolumn{1}{c}{FB}  & \multicolumn{2}{c}{FR} & \multicolumn{2}{c}{CO} & \multicolumn{2}{c|}{YN}\\ \cline{2-8}
         &Consistency&Consistency&Contrast&Consistency& Contrast&Consistency& Contrast\\ \hline
         \multirow{1}{*}{BERT}   &69.44  &76.13  &42.47  &16.99  &15.58  &48.07   &71.41\\ \hline
         \multirow{1}{*}{AlBERT} &84.77  &82.42  &41.69  &58.42  &62.51  &48.78   &69.19\\  \hline
         \multirow{1}{*}{XLNet}  &85.2   &88.56  &50     &71.10 & 72.31  &51.08   &69.18\\\hline
    \end{tabular}
    }
    \caption{Evaluation of consistency and semantic sensitivity of models in Table~\ref{tab:evaluation}. All the results are on the correctly predicted questions of {\em Seen} test set of \auto{}.}
    \label{tab:extra_evaluation}
\end{table*}

\subsection{\textsc{SpartQA} is challenging}

In addition to BERT, we continue to test another 
two LMs, ALBERT and XLNet (Table \ref{tab:extra_evaluation}). We further pretrain these LMs on \auto{}, and test them on \human{} 
% \rk{for \human we also fine-tune the models on \human training examples. We should mention this here too.}
(the numbers of BERT are copied from Table~\ref{tab:experiments}) and two held-out test sets of \auto{}, {\em Seen} and {\em Unseen}. 
Note that when a system is tested against \human{}, it is fine-tuned on \human{}'s training data following its further pretraining on \auto{}.
We use the unseen set to test to what extent the baseline models use shortcuts in the language surface. This set applies minor modifications randomly on a number of stories and questions to change the names of shapes, colors, sizes, and relationships in the vocabulary of the stories, which do not influence the reasoning steps (more details in Appendix~\ref{sec:unseen}).

% \paragraph{Baseline LMs on \auto{}:} In a different set of experiments, we investigate whether \dataset{} is generally challenging to be solved (see Table \ref{tab:evaluation}). Thus, we use three different language models BERT~\cite{devlin-etal-2019-bert}, ALBERT~\cite{lan2019albert}, and XLNet~\cite{yang2019xlnet} in the same architecture described in Sec. \ref{Sec:Architecture} and evaluate their performance on two automatically generated evaluation sets, seen and unseen sets, plus the \human{} test set. To evaluate the models performance on the human test set, we further fine-tune models on \human{} training samples.

% We use the unseen set to \qn{test to what extent} the baseline models use shortcuts in the language surface. This set \qn{applies minor modifications randomly on a number of stories and questions to change the names of shapes, colors, sizes, and relationships in the vocabulary of the stories, which do} not influence the reasoning steps (more details in Appendix~\ref{sec:additional_eval}).
% and applies minor modifications in the vocabulary of the stories. 
% The modifications are randomly applied on a number of generated stories and questions to change the names of shapes, colors, sizes, and relationships~(more details in Appendix~\ref{sec:additional_eval}).

All models perform worst in YN across all \qtype{}s, which suggests that YN presents a more complex phenomena, probably due to additional %using 
quantifiers in the questions. XLNet performs the best on all \qtype{}s except its accuracy on \human{}'s YN section. However, the drops in {\em Unseen} and {\em human} suggest overfitting %shows that the model is fitted 
on the training vocabulary. % and cannot generalize well. 
The low accuracies on human test set from all models show that solving this benchmark is still a challenging problem and requires more sophisticated methods like considering spatial roles and relations extraction~\cite{kordjamshidi2010spatial,dan2020spatial, rahgooy2018visually} to understand stories and questions better.

% The human performance is evaluated on 100 random questions from \auto{} and \human{} test sets. The respondents are graduate students that were trained by some samples of the dataset before answering the final questions.

% \subsection{Results}
% \rk{YN f1 measure in another table?}\\
% \rk{incomplete}
% \rk{Consistency and contrast part}\\
% \outline{ EXP1-5, selective masking model, test on bAbI and boolQ} \outline{remove 8K results and just mention them as an experiment}\\
% In this section, we compare the results of the experiments introduced in Section~\ref{Sec:Setup}. 

% \pk{this paragraph needs revision, currently it is just confusing, why you bring the calculation here? you even do not mention the results. You can shortly say that "we have done these two tests [cite the reference]" these are harder tests that not only test the correctness but also the same questions should get a similar answer ... [short description of consistency and contrast go here]. The results show lower performance when considering these two metrics which means the LMs remember the language surface and do not necessarily perform an actual spatial reasoning. [Remove 84*80... } 
To evaluate the reliability of the models, we also provide two extra consistency and contrast test sets. 
\textbf{Consistency set} is made by changing a part of the question in a way that seeks for the same information~\cite{hudson2019gqa, suhr2018corpus}. Given a pivot question and answer of a specific consistency set, answering other questions in the set does not need extra reasoning over the story.

\textbf{Contrast set} is made by minimal modification in a question to change its answer~\cite{gardner2020evaluating}.  For contrast sets,  there is a need to go back to the story to find the new answer for the question's minor variations~(see Appendix \ref{sec:cons} for examples.) 
% \qn{I don't understand this}\rk{it means that to solve them we should reason over new information from data. not the same information that we used in questions.} 
The consistency and contrast sets are evaluated only on the correctly predicted questions to check if the actual understanding and reasoning occurs.
This ensures the reliability of the models. %whether the model is reliable on detecting the objects and reasoning about the relations when answering those questions.

Table~\ref{tab:extra_evaluation} shows the result of this evaluation on four \qtype{}s of \auto{},
where we can see, for another time, that
% . This table shows that 
the high scores %result 
on the {\em Seen} test set are likely % mainly 
due to overfitting on training data rather than correct detection of spatial terms and reasoning over them.

% \pk{both sets together? ore once for consistency and once for contrast? rephrase if it is the second case. }

% \outline{on other target tasks}
\subsection{Extrinsic evaluation}
In this subsection, we take BERT as an example to show, once pretrained on \auto{}, BERT can achieve better performance on two extrinsic evaluation datasets, namely \babi{} and boolQ.

% \paragraph{BERT$_{QA}$--{\small\auto{}} on other dataset} In this experiment, we want to check whether the knowledge learned from \auto{} is transferable to other target tasks or benchmarks. We use BERT$_{QA}$--{\small\auto{}} trained on YN \qtype{} and check its performance on two different benchmarks, bAbI and boolQ, by fine-tuning on their training set.
% These experiments use the trained model on the YN \qtype{} and further fine-tune it on bAbI and boolQ.
% We set two other experiments to check how we can transfer the learning from \auto{} to solve other datasets' problem.

\begin{figure}[h]
    \centering
    \includegraphics[width=\linewidth]{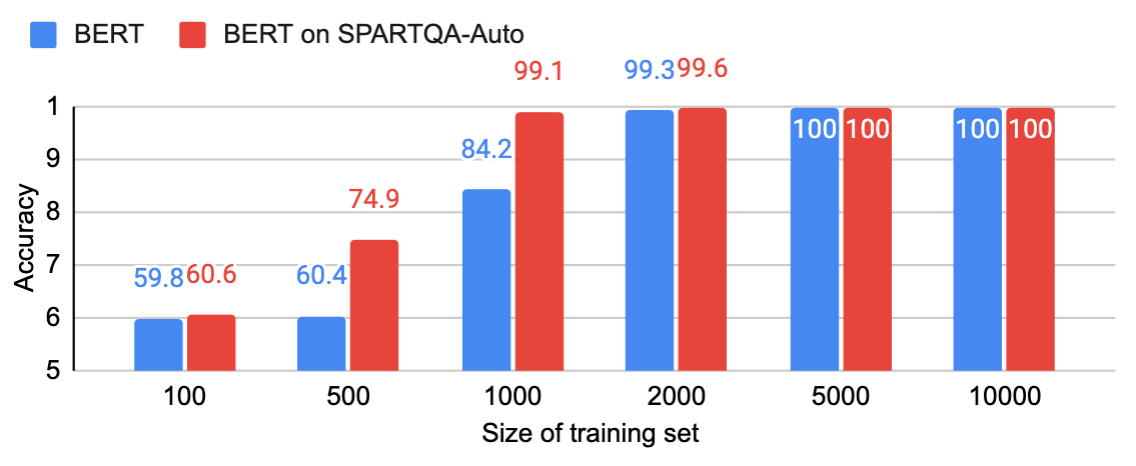}
    \caption{Learning curve of BERT and BERT further pretrained on \auto{} on \babi{}.
    % Results of  BERT and Fine-tuned BERT on \auto{} on different portion of training set of bAbI.
    }
    \label{fig:babi_train}
\end{figure}
% boolq
\begin{table}[h]
    \centering
    % \resizebox{\linewidth}{!}{%
    \begin{tabular}{|l|c|}
        \hline
        \textbf{Model} &\textbf{Accuracy}\\ %\cline{3-14} 
        \hline
          Majority baseline       & 62.2 \\
          Recurrent model (ReM)        & 62.2 \\
          ReM fine-tuned on SQuAD  & 69.8 \\
          ReM fine-tuned on QNLI   & 71.4 \\
          ReM fine-tuned on NQ  & 72.8 \\\hline
           BERT (our setup)  & 71.9 \\
          BERT (\auto{}) & \textbf{74.2} \\\hline
    \end{tabular}
    %  }
    \caption{System performances on the dev set of \boolq{} (since the test set is not available to us). Top: numbers reported in \cite{clark2019boolq}. Bottom: numbers from our experiments. BERT (\auto): further pretraining BERT on \auto{} as a QA task.
    % Result of fine-tuned version of Recurrent Model and BERT on different dataset. All these results are on the DEV set of boolQ since the test set is not provided yet. *: All models and results are from \cite{clark2019boolq}.
    } 
    \label{tab:boolq}
\end{table}

We draw the learning curve on \babi{}, using the original BERT as a baseline and BERT further pretrained on \auto{} (Fig.~\ref{fig:babi_train}).
Although both systems achieve perfect accuracy given large enough training data (i.e., 5k and 10k), BERT (\auto{}) is showing better scores given less training data. Specifically, to achieve an accuracy of 99\%, BERT (\auto) requires 1k training examples, while BERT requires twice as much.
We also notice that BERT (\auto) converges faster in our experiments.
%  \qn{I find it not very strict to mention your num of epochs without further details, so maybe just tone down a little bit and say we observe it is generally faster.}

% Figure \ref{fig:babi_train} compares the performance of BERT and BERT$_{QA}$-{\small\auto{}} on different portions of bAbI training set. Figure \ref{fig:babi_train} shows that BERT$_{QA}$-{\small\auto{}} performs better than BERT when reducing the number of bAbI training samples to 1k, 500, and 100. With 5k and 10k training examples, %samples, 
% both models can achieve 100\% accuracy while fine-tuned BERT on \auto{} achieves this performance requiring fewer training epochs (BERT$_{QA}$-{\small\auto{}} achieves 100\% accuracy with 10k training samples, after two epochs while BERT achieves this performance after 6 epochs). In conclusion, using \auto{} reduces the need of large training data and helps models converge quicker.
% \pk{clarify the point of this graph in one clear sentence, something like: Using SPRATQ no need to train on babi to solve the babi task or "Using spartqa smaller number of babi training examples are needed for solving babi task and the models converges quicker" ...}

As another evaluation dataset, we chose boolQ for two reasons. First, we needed a QA dataset with Yes/No questions. To our knowledge boolQ is the only available  one used in the recent  work.  
Second, indeed, \dataset{} and boolQ are from different domains, however, boolQ needs multi-step reasoning in which we wanted to see if \dataset{} helps. 

Table \ref{tab:boolq} shows that further pretraining BERT on \auto{} yields a better result than the original BERT and those reported numbers in \citet{clark2019boolq}, which also tested on various distant supervision signals such as SQuAD~\cite{rajpurkar2016squad}, Google's Natural Question dataset NQ~\cite{kwiatkowski2019natural}, and QNLI from GLUE~\cite{wang2018glue}. 
% This result is significant since these datasets and tasks are more related to boolQ than \dataset{}.  \qn{this statement is very loose}

% Table \ref{tab:boolq} shows that further pretraining BERT on \auto{} yields a better result than fine-tuning Recurrent model on SQuAD dataset~\cite{rajpurkar2016squad}, Natural Question dataset NQ~\cite{kwiatkowski2019natural}, and QNLI task from GLUE~\cite{wang2018glue}. This result is significant since these datasets and tasks are more related to boolQ than \dataset{}. 

% Both boolQ and SQuAD are on the same source data (Wikipedia articles), boolQ builds upon the NQ which contains some natural yes/no questions, and in QNLI task the model must determines if a SQuAD sentence contains the answer to the question or not. 

% Table \ref{tab:boolq} shows that fine-tuning BERT on our dataset yields a better result than fine-tuning Recurrent model on SQuAD dataset (that both boolQ and SQuAD~\cite{rajpurkar2016squad} are from the same source data, Wikipedia articles), Natural Question dataset , NQ~\cite{kwiatkowski2019natural} (that boolQ is built upon this) and QNLI task from GLUE~\cite{wang2018glue} (that determine if a SQuAD sentence contains the answer to the question) on the boolQ benchmark. 
% \pk{i tried to clean this paragraph, but still is grammatical, please revise. what is the "on the boolQ benchmark" at the end!}

We observe that many of the boolQ examples answered correctly by the BERT further pretrained on \auto{} require multi-step reasoning. Our hypothesis is that since solving \auto{} questions needs multi-step reasoning, fine-tuning BERT on \auto{} generally improves this capability of the base model.

\section{Conclusion}
\label{conclusion}
Spatial reasoning is an important problem in natural language understanding. We propose the first human-created QA benchmark on spatial reasoning, and experiments show that state-of-the-art pretrained language models (LM) do not have the capability to solve this task given limited training data, while humans can solve those spatial reasoning questions reliably. To improve LMs' capability on this task, we propose to use hand-crafted grammar and spatial reasoning rules to automatically generate a large corpus of spatial descriptions and corresponding question-answer annotations; further pretraining LMs on this distant supervision dataset significantly enhances their spatial language understanding and reasoning. We also show that a spatially-improved LM can have better results on two extrinsic datasets~(\babi{} and \boolq{}). 

% The existing QA corpora do not contain complex spatial questions and answers. Hence, we provide the first QA, \human, evaluation benchmark on textual spatial reasoning. Our experiments show the SOTA pre-trained language models fine-tuned on QA
% trained on existing data 
% do not perform well on our human-generated dataset. We propose a new synthetic QA dataset, \auto, using hand-crafted grammar and spatial rules as a source of distant supervision and generate a large corpus of spatial descriptions and complex spatial questions.
% We show that training models on \auto{} enhances their spatial language understanding and reasoning.

% We also show that a fine-tuned language model on \auto{} can have better results on other datasets~(\babi{} and \boolq{}).
% This data is created based on spatial semantics based on the geometrical properties on objects and rules of spatial reasoning. As a future direction, we will investigate the spatial reasoning based on functional properties of objects and relationships rather than geometric ones; this will need further commonsense knowledge about the functionality of the relationships and object affordances.

\section*{Acknowledgements}

This project is supported by National Science Foundation (NSF) CAREER award \texttt{\#}2028626 and (partially) supported by the Office of Naval Research grant \texttt{\#}N00014-20-1-2005. 
We thank the reviewers for their helpful comments to improve this paper and Timothy Moran for his help in the human data generation. 

\bibliography{anthology,custom}
\bibliographystyle{acl_natbib}

\clearpage
\newpage
\appendix
%!TEX root=naacl2021.tex
\section{Question Templates and statistics Information}
\label{sec:supplemental}

\begin{table*}[htbp]
\centering
\begin{tabular}{|l|l|l|}
\hline
\textbf{Q-Type} & \textbf{Q-Templates}& \textbf{Candidate answer}\\ \hline
FR& what is the relation between \textless object\textgreater and \textless object\textgreater?& \begin{tabular}[c]{@{}l@{}}
     Left, Right, Below,\\Above, Touching,\\ Far from, Near to
\end{tabular}\\ \hline

CO& \begin{tabular}[c]{@{}l@{}}What is \textless relation \textgreater the \textless object\textgreater?\\\hspace{20 mm}an \textless object1\textgreater or an \textless object2\textgreater?\\ Which object is \textless relation \textgreater an \textless object\textgreater? \\\hspace{20 mm}the \textless object1\textgreater or the \textless object2\textgreater?\end{tabular}& \begin{tabular}[c]{@{}l@{}}Object1, object2,\\ Both, None\end{tabular} \\\hline

YN& \begin{tabular}[c]{@{}l@{}}Is (the $\mid$ a )\textless object1\textgreater \textless relation\textgreater (the $\mid$ a) \textless object2\textgreater?\\ Is there any \textless object1\textgreater s \textless relation\textgreater all \textless object2\textgreater s?\end{tabular}& \multicolumn{1}{l|}{Yes, No, Don't Know}\\ \hline

FB& \begin{tabular}[c]{@{}l@{}}Which block has an \textless object\textgreater? \\ Which block doesn't have an \textless object\textgreater? \end{tabular} & Name of blocks, None\\ \hline

% FA& What is the \textless attribute\textgreater of \textless object\textgreater?& Span of the text\\ \hline
\end{tabular}%
\caption{Questions and answers templates.}
\label{tab:templates}
\end{table*}

Table~\ref{tab:templates} shows the templates used to create questions in \auto{}. The ``\textless object\textgreater'' is a variable replaced by objects from the story (using \textit{Choose-objects} and \textit{Describe-objects} modules), and the ``\textless relation\textgreater'' variable can be replaced by the chosen relations between objects (using \textit{Find-all-relations} module).

The articles and the indefinite pronouns in each template play an essential role in understanding the question's objective. For example, ``Are all blue circles near to a triangle?'' is different from ``Are there any blue circles near to a triangle?'', and ``Are there any blue circles near to all triangles?''. Therefore, we check the uniqueness of the object definition, using ``a'' or ``the'' in proper places and randomly place the terms ``any'' or ``all'' in the YN questions to generate different questions.

\begin{table}[h]
\centering
\begin{tabular}{|l|l|ll|}
\hline
\qtype{} & Candidate Answers & train & test \\ \hline \hline
\multirow{8}{*}{\begin{tabular}[c]{@{}l@{}}FR\\ (Multiple\\ Choices)\end{tabular}} & Left &  20.7  & 17.9 \\ \cline{2-4} 
 & Right & 21.4 &  16.7\\ \cline{2-4} 
 & Above &  26.9& 25.4 \\ \cline{2-4} 
 & Below &  37.2& 42.9 \\ \cline{2-4} 
 & Near to &  5.8& 2.9 \\ \cline{2-4} 
 & Far from & 1.3 & 0.56 \\ \cline{2-4} 
 & Touching & 0.57 & 0.27 \\ \cline{2-4} 
 & DK & 0.52 & 0.32 \\ \hline \hline
\multirow{4}{*}{\begin{tabular}[c]{@{}l@{}}FB\\ (multiple\\ Choices)\end{tabular}} & A & 49.8  &  49.4\\ \cline{2-4} 
 & B & 50.1 &  50\\ \cline{2-4} 
 & C &  35.1& 62 \\ \cline{2-4} 
 & {[}{]} & 7.1 &  90.5\\ \hline \hline
\multirow{5}{*}{\begin{tabular}[c]{@{}l@{}}CO\\ (Single \\ choice)\end{tabular}} & Object1 & 25.4 &  26\\ \cline{2-4} 
 & Object2 & 25.3 &  24.9\\ \cline{2-4} 
 & Both & 44.3 & 43.9 \\ \cline{2-4} 
 & None & 4.9 & 5.0 \\ \hline \hline
\multirow{3}{*}{\begin{tabular}[c]{@{}l@{}}YN\\ (Single\\ choice)\end{tabular}} & Yes &  53.3& 50.5 \\ \cline{2-4} 
 & No & 18.7 & 23.6 \\ \cline{2-4} 
 & DK & 27.8 & 25.9 \\ \hline
\end{tabular}
\caption{The percentage of each correct label in all samples. *The candidate answers for the FB \qtype{} can be varied, based on its story. **CO can be considered as a multiple choice or single choice question. E.g., in "which object is above the triangle? the blue circle or the black circle?" you can consider two labels with boolean classification on each "blue circle" and "black circle" or consider it as a four labels classification: "blue circle," "black circle," "both of them," and "None of them." *** \textbf{DK}, \textbf{None}, \textbf{[]}, all mean none of the actual labels are correct.}
\label{tab:number_of_choices}
\end{table}

Table \ref{tab:number_of_choices} shows the percentage of correct labels in train and test sets. In multi-choice \qtype s, more than one label can be true. 
% In single choice \qtype s, for each question only one label can be chosen, so the summation of the correct labels are equal to one, while 
% In multiple choices \qtype s, each label (except DK) has equal chance to be chosen. For instance, this table shows in 49.4\% of FB questions in test set, label A (block A) is one of the answers.

% Table \ref{tab:number_of_choices} shows \pk{the statistics are the correct percentage of each label in the whole samples:?}. In single choices, \qtype s, the summation of the labels' percentage are equal to one while for the other, \pk{it is independent for each label: the labels are independent from each other? }. It means that the label "left" is one of the correct answers in all samples\pk{this is not understandable for me, are you giving an example? why left?}.

\section{Sentences of the Dataset}
Table~\ref{tab:particular_features} shows some generated sentences in \auto{} with some specific features that challenge models to understand different forms of relation description in spatial language. 

\section{Additional Evaluation Sets}
\label{sec:additional_eval}
Here we describe three extra evaluation sets provided with this dataset in more detail, including unseen test, consistency, and contrast sets.

\subsection{Unseen Evaluation Set}
\label{sec:unseen}
We propose an unseen test set alongside the seen test of \auto{} to check whether a model is using shortcuts in the language surface by describing objects and relations with new vocabularies in the samples. This set has minor modifications that should not affect the performance of a consistent and reliable model. The modifications are randomly applied on a number of generated stories and questions and include changing names of shapes, colors, sizes, and relationships' names ~(describing relationships using different language expressions). The modification choices are described in Table~\ref{tab:modifications}.

\begin{table}[h]
    \centering
    \begin{tabular}{l l l}
    \hline
         \textbf{Type}& \textbf{Original Set}& \textbf{Unseen Set} \\ \hline
         Shapes&\begin{tabular}[c]{@{}l@{}} Square, Circle, \\ Triangle \end{tabular}& \begin{tabular}[c]{@{}l@{}} Rectangle, Oval, \\Diamond \end{tabular}\\ \hline
         Relations &\begin{tabular}[c]{@{}l@{}} Left, Right,\\ Above, Below \end{tabular}& \begin{tabular}[c]{@{}l@{}} Left side,\\ Right side, \\Top, Under \end{tabular}\\ \hline
         Colors&\begin{tabular}[c]{@{}l@{}} Yellow, Black,\\ Below \end{tabular}& \begin{tabular}[c]{@{}l@{}} Green, Red,\\ White \end{tabular}\\ \hline
         Size&\begin{tabular}[c]{@{}l@{}} Small,\\ Medium, Big \end{tabular}& \begin{tabular}[c]{@{}l@{}} Little, Midsize,\\ Large \end{tabular}\\
         \hline
    \end{tabular}
    \caption{Modifications on the unseen set}
    \label{tab:modifications}
\end{table}

\subsection{Contrast and Consistency Evaluation}
\label{sec:cons}
For probing the consistency and semantic sensitivity of models, we provide two extra evaluation test sets, Consistency and Contrast\footnote{for some questions, it is not possible to generate a complementary set}.

\textbf{Consistency set} is made by changing parts of the question in a way that it still asks about the same information~\cite{hudson2019gqa, suhr2018corpus}. For instance, for the question, ``What is the relation between the blue circle and the big shape? Left,'' we create a similar question in the form of ``What is the relation between the big shape and the blue circle? Right''. Answering these questions around a pivot question is possible for human without the need for extra reasoning over the story and based on the main questions' answer. Hence, the evaluation on this set shows that models understand the real underlying semantics rather than overfit on the structure of questions. 
% answering \pk{based on an unrelated scheme in the data:?}.

 \textbf{Contrast set}: This set is made by minor changes in a question that changes the answer~\cite{gardner2020evaluating}. As an instance, in the question ``Is the blue circle below the black triangle? Yes,'' we create a contrast question ``Is the blue circle below all triangles? No'' by changing ``the black trinagle'' to ``all triangles''. The evaluation on this set shows the robustness of the model and its sensitivity to the semantic changes when there are minor changes in the language surface~\footnote{Based on the original contrast set paper, consistency and contrast set should be generated manually to control the semantic change. In our case that we are probing the spatial language understanding of models, we must change parts that affect spatial understanding, which can be implemented by some static rules.}.

\section{Extra Annotations}
Alongside the main \auto's stories and questions we provided some extra annotation to help the models to understand the spatial language better. 
% to be able to use in more sophisticated models 
% \pk{to understand the text better.:?}

\subsection{Detailed Annotation and Scene-Graphs}
\label{sec:annotation}
\begin{figure}
    \centering
    \includegraphics[width= \linewidth]{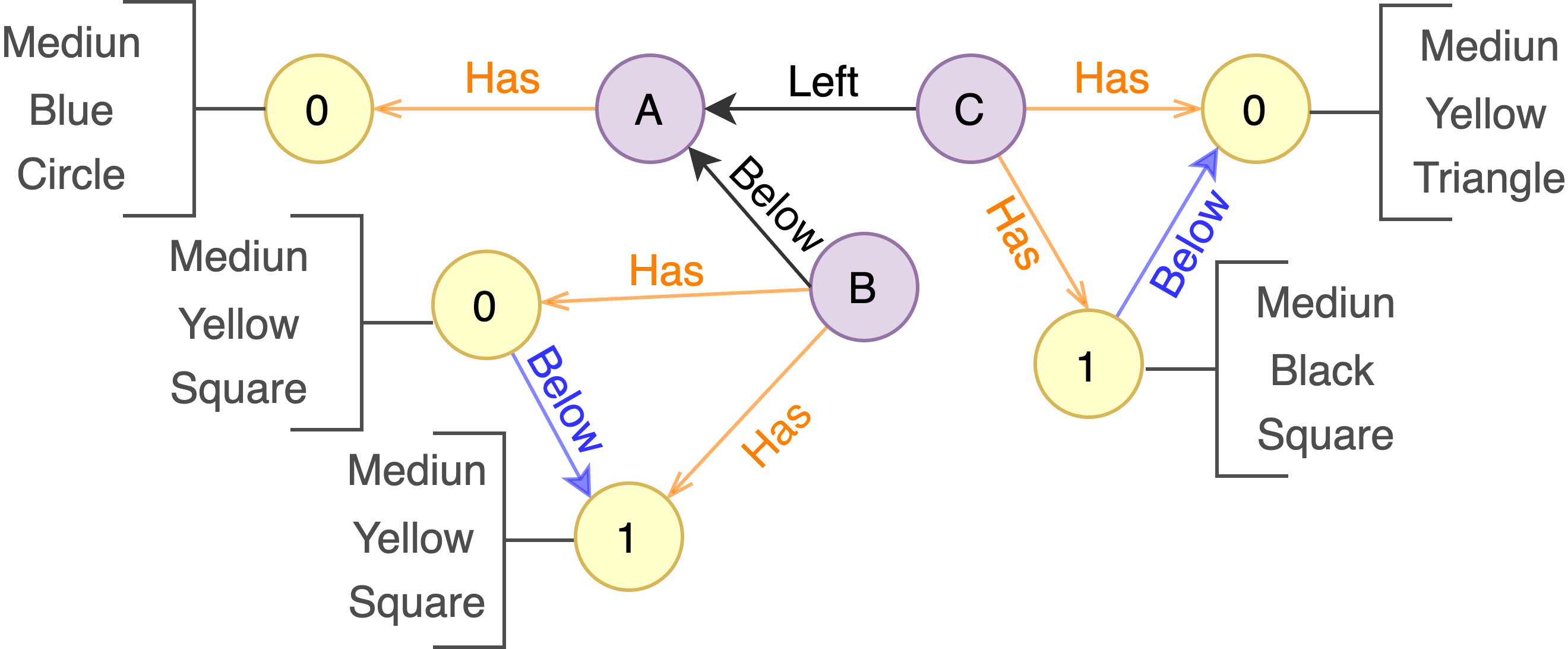}
    \caption{Scene-graph}
    \label{fig:scene-graph}
\end{figure}

Providing in-depth human annotations is quite expensive and time-consuming. 
In \auto{}, we generated fine-grained scene-graph based on the story. This scene-graph contains blocks' description, their relations, and the objects' attributes alongside their direct relations with each other.
The scene-graphs can be used for the models to understand all spatial relations directly mentioned in the textual context. Figure~\ref{fig:scene-graph} shows an example of this scene-graph. The scene-graph can provide strong supervision for question answering challenges and can be used to evaluate models based on their steps of reasoning and decisions.

\subsection{SpRL Annotation}
\label{sec:sprlAnnotation}
\begin{figure*}[b]
    \centering
    \includegraphics[width=\linewidth]{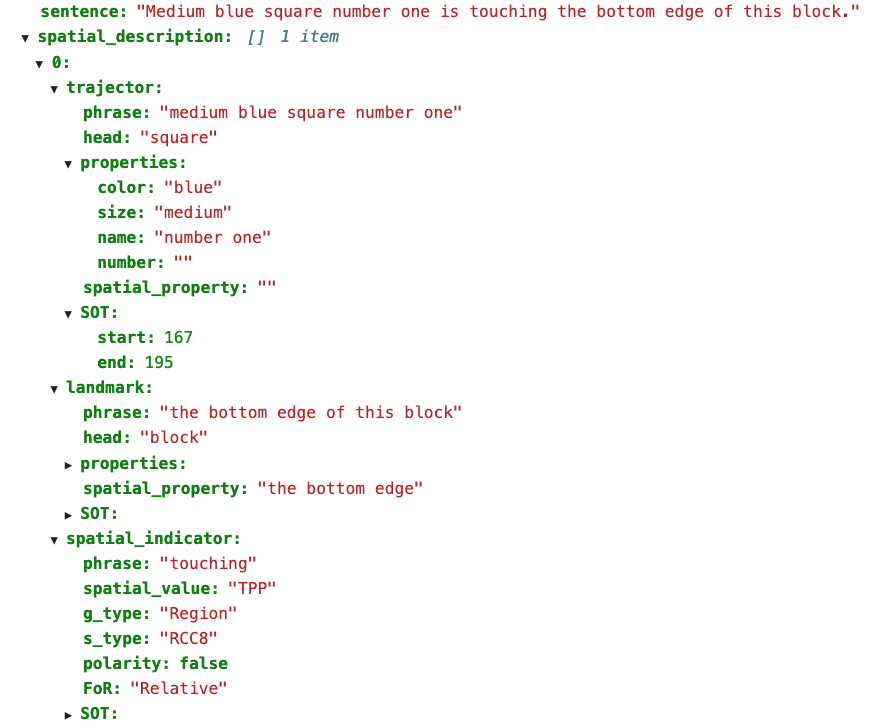}
    \caption{SpRL annotation for an example sentence from \dataset.}
    \label{fig:sprl}
\end{figure*}

We also provided spatial annotations for each sentence and question, based on Spatial Role Labeling (SpRL) annotation scheme~\cite{kordjamshidi2010spatial}(Fig. \ref{fig:sprl}). This annotation is generated by hand-crafted rules during the main data generation. SpRL is used for recognizing spatial expressions and arguments in a sentence. This annotation is useful for applications that need to detect and reason about spatial expressions and arguments.

\section{QA Language Models for Spatial Reasoning over Text}
\label{sec:lm-arch}
\begin{figure}[htb!]
	\centering
	\begin{subfigure}[b]{\linewidth}
		\includegraphics[width=\linewidth]{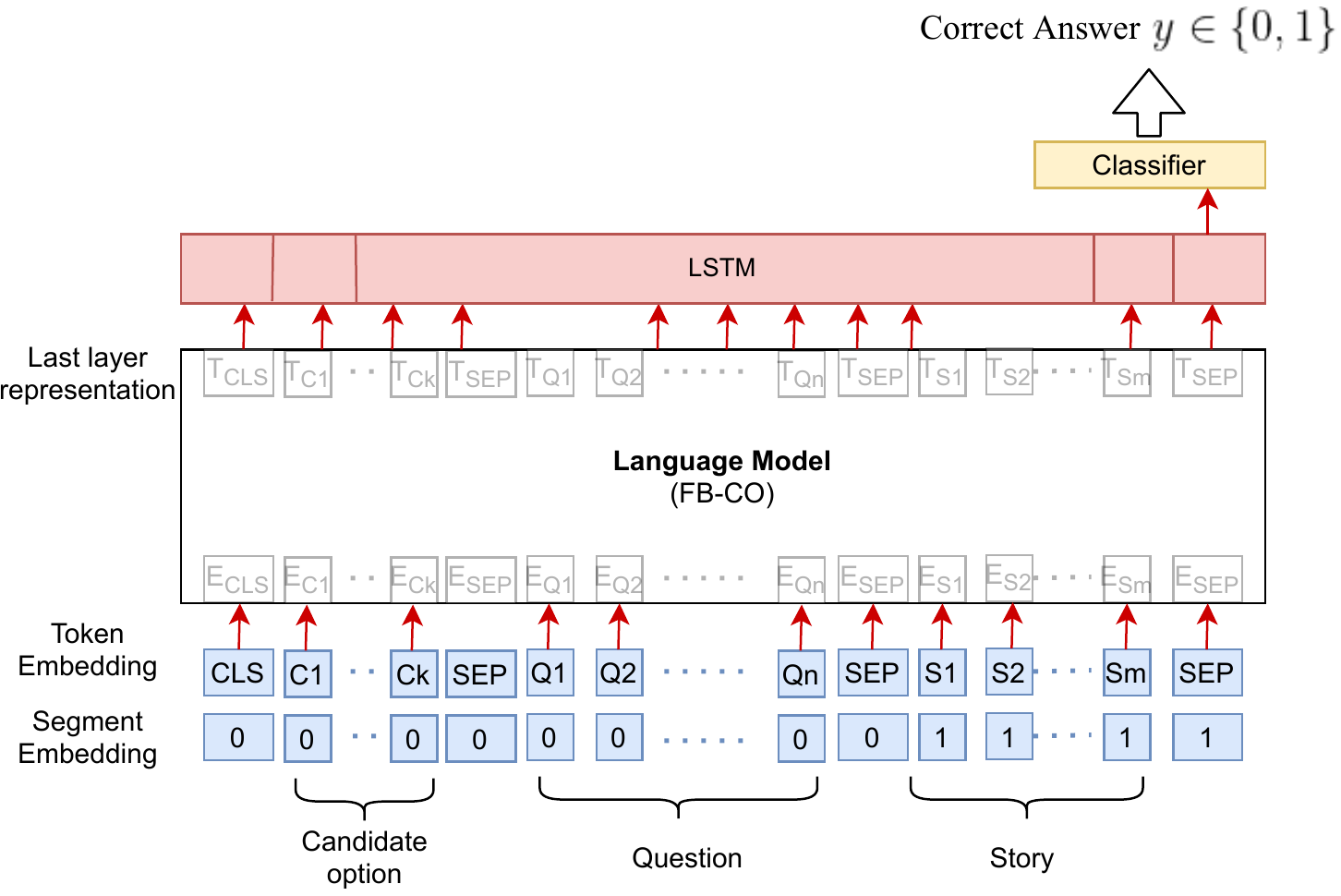}
		\caption{LM$_{QA}$ Architecture for CO and FB \qtype{}s}
		\label{fig:modelfb}
	\end{subfigure}
%	\hfill
	\begin{subfigure}[b]{\linewidth}
			\includegraphics[width=\linewidth]{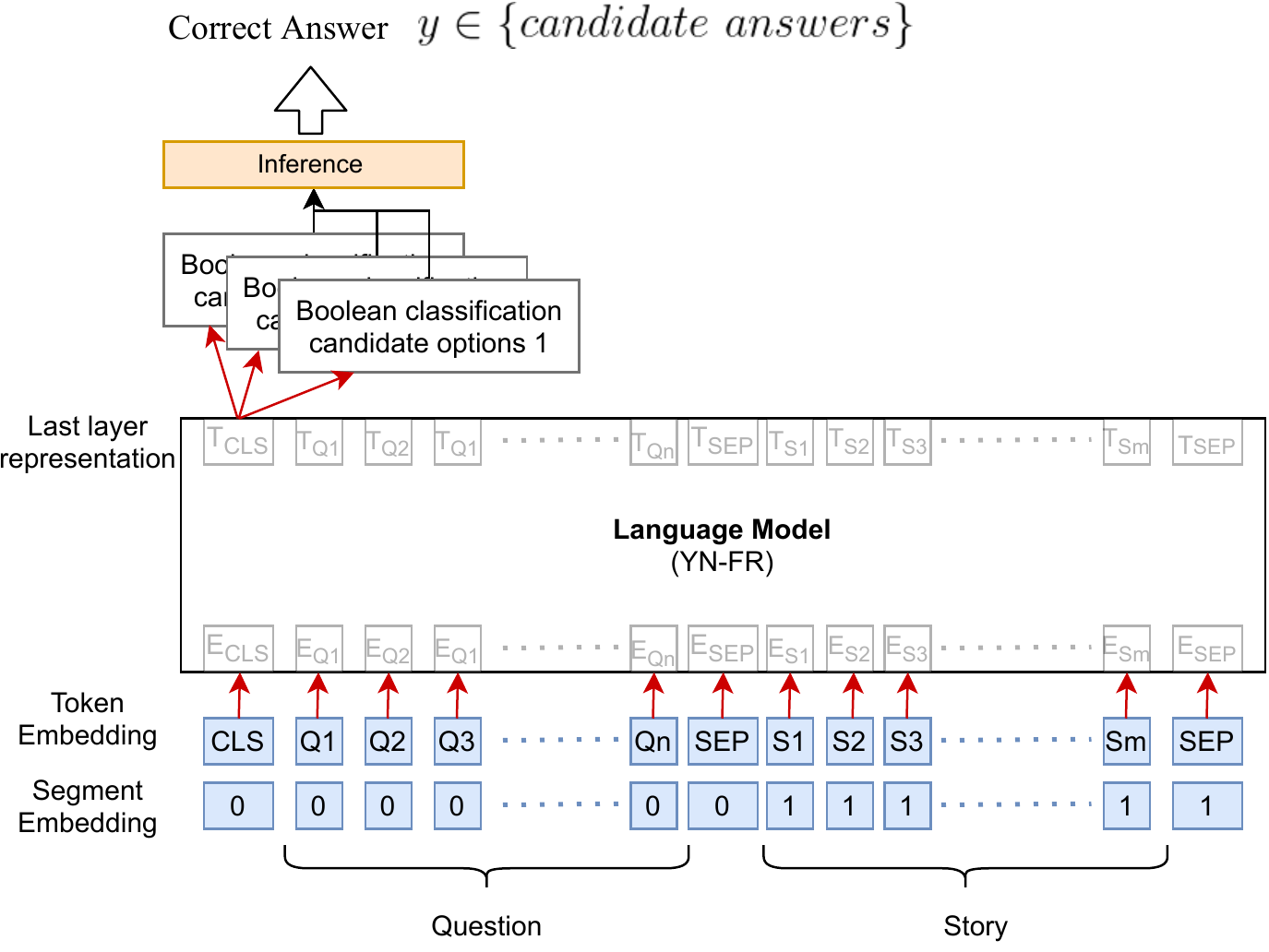}
			\caption{LM$_{QA}$ Architecture for FR and YN \qtype{}s}
			\label{fig:modelyn}
    \end{subfigure}
 	\caption{LM$_{QA}$ for Spatial Reasoning over Text}
	\label{fig:model}
\end{figure}

Figures \ref{fig:modelfb} and \ref{fig:modelyn} depict the architecture used for further fine-tuning language models on \dataset{} described in section 5.

\section{\babi{} and \boolq{} Datasets}
\label{sec:babiboolq}
\begin{figure}[htb!]
    \centering
    \includegraphics[width=\linewidth]{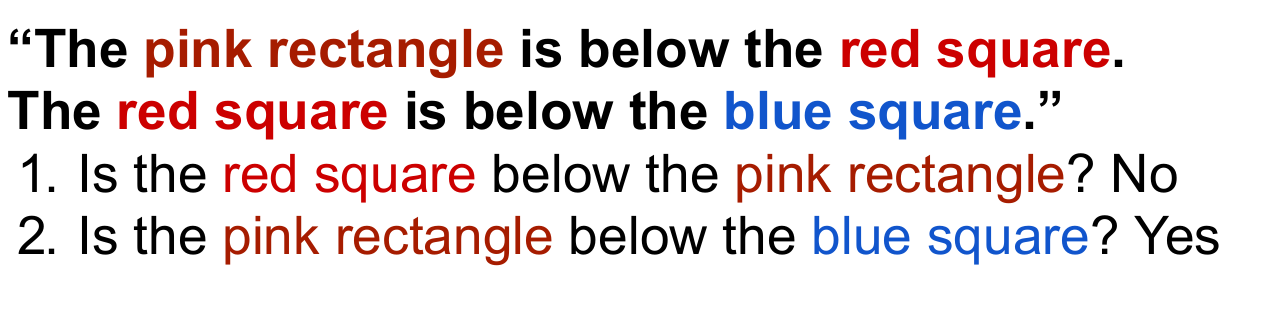}
    \caption{An example of bAbI dataset, task 17.}
    \label{fig:babi}
\end{figure}

Figure~\ref{fig:babi} shows an example of the \babi{} dataset~\cite{weston2015towards} task 17. 
% The story is quite simple and answering the questions do not require more than 2 steps of reasoning. 

To solve task 17 of \babi{} , we implement two SpRL+rule-based and neural network models.
The SpRL+rule-based model first, finds different spatial relation triplets (Landmark, Spatial-indicator, trajector) for each fact in a story the applies spatial rules over these extracted triplets and report all possible relations between two asked objects. Finally, it checks whether the asked relation existed in the find relation. This model solves task 17 of the \babi{} with $100\%$ accuracy.

% an inference algorithm based on relation properties on extracted triplets. Finally, we check if the question relation triplets can be verified using the inference algorithm. This model solves task 17 of the bAbI dataset with $100\%$ accuracy.

To implement the neural network approach, we use huggingface implementation of pre-trained BERT~\cite{devlin-etal-2019-bert}. We apply a boolean classifier on the output of ``[CLS]'' token from the last layer of BERT model for each ``Yes'' and ``No'' answers (the same as model used on YN question types.) We use Adamw~\cite{loshchilov2017decoupled} optimizer and $2e-6$ learning rate with negative log-likelihood loss objective and train the model on the 10k, 5k, 2k, 1k, 500, and 100 portion of \babi's training questions. The model yields $100\%$ accuracy on 10k, and 5k and $99\%$ accuracy on 2k and 1k training samples.

\begin{figure}[h]
    \centering
    \includegraphics[width=\linewidth]{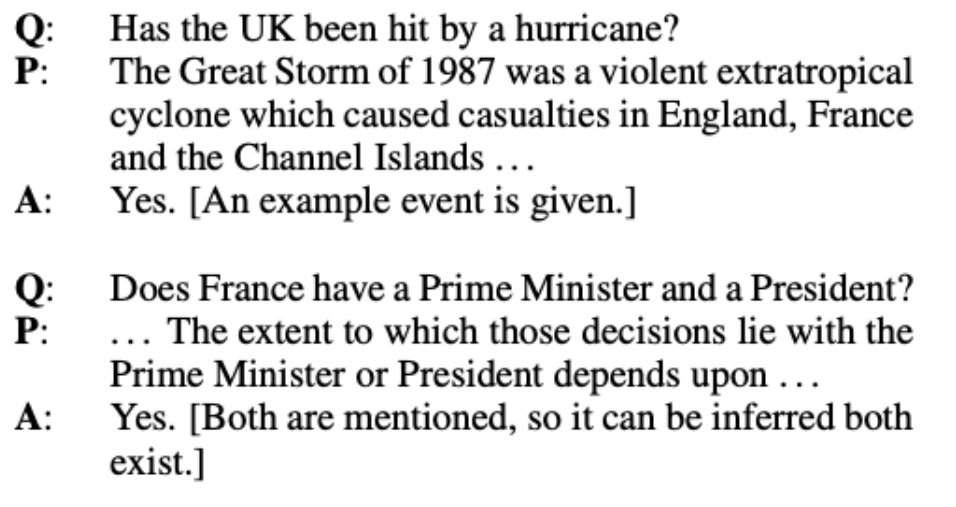}
    \caption{An example of boolQ dataset.}
    \label{fig:boolq}
\end{figure}

Figure \ref{fig:boolq} shows an example of \boolq{} dataset. To Answering the questions of this dataset, we use the same setting as neural network model on \babi{} to further fine-tune BERT on \boolq{}.

\begin{table*}[b]
\centering
\begin{tabular}{p{0.45\linewidth}|p{0.45\linewidth}}
\hline
\textbf{Examples} & \textbf{Features} \\ \hline
Block A is above Block C \textbf{and} B. & Using conjunction to describe relation between more than two blocks. \\ \hline
The small circle is \textbf{above} the yellow square \textbf{and} the big black shape. & Using conjunction to describe relationships between more than two objects. \\ \hline
The yellow square number one is to the \textbf{right} of \textbf{and} \textbf{above} the blue circle. & Using conjunction for more than one relation. \\ \hline
Block B has \textbf{two medium yellow squares} and \textbf{two blue circles}. & Describing a group of objects with the same properties. In the next sentences, they are mentioned by an asigned number. For example, the blue circle number two. \\ \hline
The blue circle is below the object\textbf{ which is to the right} of the big square. & Using nested relations between objects in their description. \\ \hline
% \textit{One medium yellow triangle and a small yellow square are in a block.} & There is another block to the left of the first block. It contains a small blue circle. \\ \hline

A small blue circle is near to the big circle. \textbf{It} is to the left of the medium yellow square. & Using coreferences for an entity described in the previous sentences. \\ \hline

There \textbf{is a} block named A. One small yellow square \textbf{is} touching the bottom edge of this block. & The verb matches the number of the subject. 
% A relation described which is the combination of RCC-8 and directions. 
\\ \hline

What is the relation between black \textbf{object} and a big circle? & Using shape, object, and thing, which are a general description of an object. It could be the ``black triangle'' or the ``black circle'' mentioned in the story.\\ \hline

\end{tabular}%
\caption{Particular features of the dataset}
\label{tab:particular_features}
\end{table*}

\end{document}